\documentclass[10pt,twocolumn,letterpaper]{article}

\usepackage{cvpr}              

\usepackage{graphicx}
\usepackage{amsmath}
\usepackage{amssymb}
\usepackage{booktabs}
\usepackage{times}
\usepackage{epsfig}
\usepackage{array}
\usepackage{makecell}
\usepackage{multirow}
\usepackage{animate}
\usepackage{wrapfig}
\usepackage{enumitem}
\usepackage{float}
\usepackage[misc]{ifsym}

\newcolumntype{P}[1]{>{\centering\arraybackslash}p{#1}}

\usepackage[pagebackref,breaklinks,colorlinks]{hyperref}

\makeatletter
\def\ps@myheadings{%
    \let\@oddfoot\@empty\let\@evenfoot\@empty
    \def\@evenhead{\thepage\hfil\slshape\leftmark}%
    \def\@oddhead{{\slshape\rightmark}\hfil\thepage}%
    \let\@mkboth\@gobbletwo
    \let\sectionmark\@gobble
    \let\subsectionmark\@gobble
    }
  \if@titlepage
  \renewcommand\maketitle{\begin{titlepage}%
  \let\footnotesize\small
  \let\footnoterule\relax
  \let \footnote \thanks
  \null\vfil
  \vskip 60\p@
  \begin{center}%
    {\LARGE \@title \par}%
    \vskip 3em%
    {\large
     \lineskip .75em%
      \begin{tabular}[t]{c}%
        \@author
      \end{tabular}\par}%
      \vskip 1.5em%
    {\large \@date \par}
  \end{center}\par
  \@thanks
  \vfil\null
  \end{titlepage}%
  \setcounter{footnote}{0}%
}
\else
\renewcommand\maketitle{\par
  \begingroup
    \renewcommand\thefootnote{\@fnsymbol\c@footnote}%
    \def\@makefnmark{\rlap{\@textsuperscript{\normalfont\@thefnmark}}}%
    \long\def\@makefntext##1{\parindent 1em\noindent
            \hb@xt@1.8em{%
                \hss\@textsuperscript{\normalfont\@thefnmark}}##1}%
    \if@twocolumn
      \ifnum \col@number=\@ne
        \@maketitle
      \else
        \twocolumn[\@maketitle]%
      \fi
    \else
      \newpage
      \global\@topnum\z@   
      \@maketitle
    \fi
    \thispagestyle{plain}\@thanks
  \endgroup
  \setcounter{footnote}{0}%
}
\makeatother

\usepackage[capitalize]{cleveref}
\crefname{section}{Sec.}{Secs.}
\Crefname{section}{Section}{Sections}
\Crefname{table}{Table}{Tables}
\crefname{table}{Tab.}{Tabs.}

\newcounter{alphasect}
\def\alphainsection{0}

\let\oldsection=\section
\def\section{%
  \ifnum\alphainsection=1%
    \addtocounter{alphasect}{1}
  \fi%
\oldsection}%

\renewcommand\thesection{%
  \ifnum\alphainsection=1%
    \Alph{alphasect}
  \else%
    \arabic{section}
  \fi%
}%

\newenvironment{alphasection}{%
  \ifnum\alphainsection=1%
    \errhelp={Let other blocks end at the beginning of the next block.}
    \errmessage{Nested Alpha section not allowed}
  \fi%
  \setcounter{alphasect}{0}
  \def\alphainsection{1}
}{%
  \setcounter{alphasect}{0}
  \def\alphainsection{0}
}%

\def\cvprPaperID{3579} 
\def\confName{CVPR}
\def\confYear{2022}

\begin{document}

\title{Neural Texture Extraction and Distribution for  Controllable \\ Person Image Synthesis}

\author{Yurui Ren$^{1}$~~~~~Xiaoqing Fan$^{1}$~~~~~{Ge Li \footnotesize{\Letter}}$^{1}$~~~~~Shan Liu$^{2}$~~~~Thomas H. Li$^{3,1}$\\
$^1$School of Electronic and Computer Engineering, Peking University~~~$^2$Tencent America~~~\\
$^3$Advanced Institute of Information Technology, Peking University~~\\
{\tt\small ~~~yrren@pku.edu.cn~~~fanxiaoqing@stu.pku.edu.cn~~~geli@ece.pku.edu.cn}\\
{\tt\small shanl@tencent.com~~~tli@aiit.org.cn~~~}
}
\maketitle

\begin{abstract}
  We deal with the controllable person image synthesis task which aims to re-render a human from a reference image with explicit control over body pose and appearance.
  Observing that person images are highly structured, we propose to generate desired images by extracting and distributing semantic entities of reference images.
  To achieve this goal, a neural texture extraction and distribution operation based on double attention is  described. 
  This operation first extracts semantic neural textures from reference feature maps. 
  Then, it distributes the extracted neural textures according to the spatial distributions learned from target poses.
  Our model is trained to predict human images in arbitrary poses, which encourages it to extract disentangled and expressive neural textures representing the appearance of different semantic entities.
  The disentangled representation further enables explicit appearance control.
  Neural textures of different reference images can be fused to control the appearance of the interested areas.
  Experimental comparisons show the superiority of the proposed model.
  Code is available at \url{https://github.com/RenYurui/Neural-Texture-Extraction-Distribution}.
\end{abstract}

\begin{figure}[t]
\centering
\includegraphics[width=0.95\linewidth]{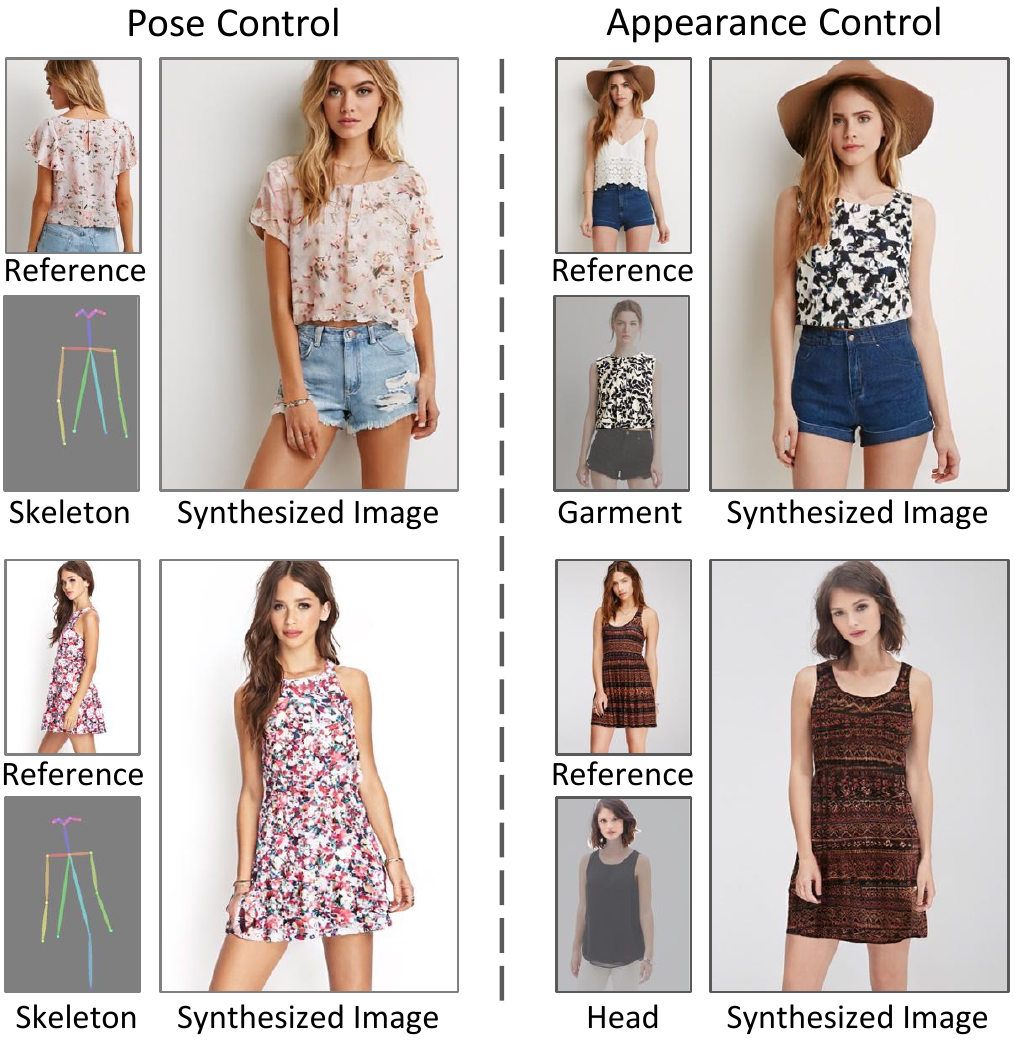}
\caption{Controllable person image synthesis. Our model can generate realistic images by explicitly controlling the poses and appearance of reference images.}
\label{fig:intro}
\end{figure}

\section{Introduction}
\label{sec:intro}
Synthesizing person images with explicitly controlling the body pose and appearance is an important task with a large variety of applications. 
Industries such as electronic commerce, virtual reality, and next-generation communication require such algorithms to generate content.
Typical examples are shown in Fig.~\ref{fig:intro}. 
It can be seen that the desired output images are not aligned with the reference images.
Therefore, a fundamental challenge for generating photo-realistic target images is to accurately deform the reference images according to the modifications.

However, Convolutional Neural Networks lack the ability to enable efficient spatial transformation~\cite{goodfellow2016deep,vaswani2017attention}. 
Building blocks of CNNs process one local neighborhood at a time. 
To model long-term dependencies, stacks of convolutional operations are required to obtain large receptive fields. 
Realistic textures will be ``washed away" during the repeating local operations.
Flow-based methods~\cite{liu2019liquid,wang2019few,siarohin2018deformable,ren2020deep} are proposed to enable efficient spatial transformation.
These methods predict 2D coordinate offsets assigning a sampling position for each target point.
Although realistic textures can be reconstructed, these methods yield noticeable artifacts, which is more evident when complex deformations and severe occlusions are observed~\cite{ren2021combining}. 

Attention mechanism~\cite{vaswani2017attention,wang2018non,zhang2019self} has emerged as an efficient  approach to capture long-term dependencies. 
This operation computes the response of a target position as a weighted sum of all source features.
Therefore, it can build dependencies by directly computing the interactions between any two positions. 
However, in this task, the vanilla attention operation suffers from some limitations. 
First, since the target images are the deformation results of the sources, each target position is only related to a local source region, which means that the attention correction matrix should be a sparse matrix to reject the irrelevant regions. Second, the quadratic memory footprint hinders its applicability to deform realistic details in high-resolution features. 

\begin{figure*}[t]
\begin{center}
\includegraphics[width=1\linewidth]{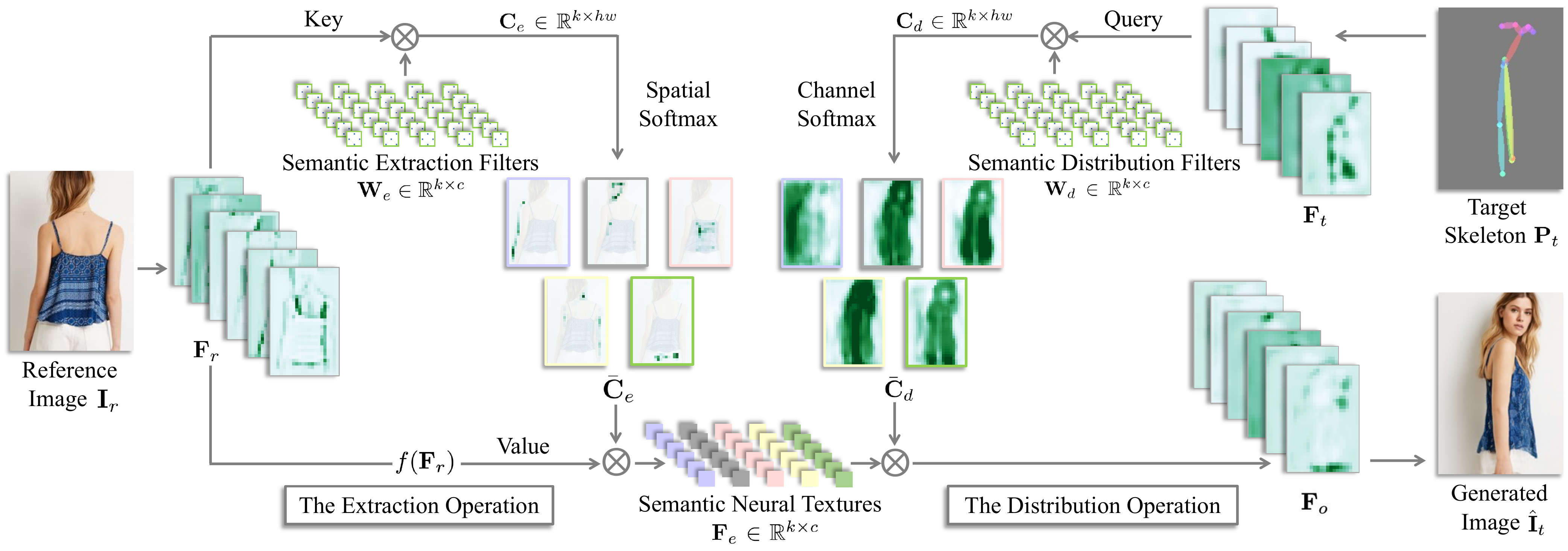}
\end{center}
   \caption{Overview of the neural texture extraction and distribution operation. Semantic neural textures are first extracted from the reference feature map. Then they are distributed according to the spatial distributions learned from the target skeleton. The heat maps show the attention coefficients $\bar{\mathbf{C}}_e$ and $\bar{\mathbf{C}}_d$. Dark color indicates high weights.}

\label{fig:hook-tile}
\end{figure*}

To deal with these limitations, we introduce an efficient spatial transformation operation.
This operation is motivated by an intuitive idea: person images can be manipulated by extracting and reassembling semantic entities (\emph{e.g.} face, hair, cloth).
To achieve this goal, we propose a Neural Texture Extraction and Distribution (NTED) operation based on double attention~\cite{chen20182,shen2021efficient}. 
The architecture of this operation is shown in Fig.~\ref{fig:hook-tile}.
Specifically, the extraction operation is first used to extract neural textures by gathering features obtained from the reference images.
Then, the distribution operation is responsible for generating the results by soft selecting the extracted neural textures  for each target position according to the learned semantic distribution.

We design a generative neural network by using NTED operations at different scales.
This network renders the input skeletons by predicting the conditional semantic distributions and reassembling the extracted neural textures.
The experimental evaluation demonstrates photo-realistic results at a high resolution of $512\times352$.
The comparison experiments show the superiority of the proposed model.
In addition, our model can be further applied for explicit appearance control.
Interested semantics can be manipulated by exchanging the corresponding neural textures of different references.
An optimization method is proposed to automatically search for the interpolation coefficients which are further used to fuse the extracted neural textures.
Our method enables coherent and realistic results.
The main contributions of our paper can be summarized as:
\vspace{-2.5mm}
\begin{itemize}
  \setlength\itemsep{-0.9mm}
  \item An intuitive idea for image deformation is provided. Desired images are generated by extracting and distributing the semantic entities of reference images.
  \item We implement the proposed idea with a light-weighted and computationally-efficient NTED operation. Experiments show the operation as an efficient spatial deformation module. Comprehensive ablation studies demonstrate its efficacy.
  \item Thanks to the disentangled and expressive neural textures extracted by our model, we can achieve explicit appearance control by interpolating between neural textures of different references.
\end{itemize}

\section{Related Work}
\noindent
\textbf{Exemplar-based Image Synthesis.} 
Recently, advances in conditional Generative Adversarial Networks~\cite{mirza2014conditional,isola2017image,wang2018high,zhu2017unpaired,zhu2017multimodal,choi2018stargan,huang2018multimodal} (cGAN) have made tremendous progress in synthesizing realistic images. 
As a typical task of cGAN, image-to-image translation~\cite{isola2017image} aims to train a model such that the conditional distribution of the generated images resembles that of the target domain. 
To achieve flexible and fine-grained control over the generated images, some exemplar-based image translation methods~\cite{huang2018multimodal,wang2018high,park2019semantic,yu2019multi} are proposed.
These methods condition the translation on an exemplar image with the desired style.
Latent vectors are extracted from exemplars to modulate the generation.
Images with specific styles are generated.
However, 1D vectors may be insufficient for representing complex textures, which hinders models to reconstruct realistic details.
Some models~\cite{zhang2020cross,Zhou_2021_CVPR} solve this problem by extracting dense semantic correspondence between cross-domain images.
The warped exemplar images provide spatially-adaptive textures, which helps with the reconstruction of local textures.

\noindent
\textbf{Pose-guided Person Image Synthesis.} 
The pose-guided person image synthesis task can be seen as a kind of exemplar-based image translation task where the appearance of the reference images is expected to be reproduced under arbitrary poses.
Some early attempts~\cite{ma2018disentangled,esser2018variational} solve this problem by extracting pose-irrelevant vectors to represent appearance. 
However, textures of different semantic entities vary greatly. 
Directly extracting vectors from reference images will limit the model to represent complex textures.
To alleviate this problem, methods are proposed to extract attributes from different segmentation regions~\cite{men2020controllable} or pre-process the reference images with UV-maps~\cite{sarkar2021style}.
These methods can extract expressive latent vectors to improve the generation quality.
However, since they apply the modulation uniformly, detailed patterns may be washed out in the final output.
To achieve spatially-adaptive modulations, dense deformations are estimated to generate aligned features by warping the references.
Flow-based methods~\cite{siarohin2018deformable,liu2019liquid,li2019dense,ren2020deep,albahar2021pose,tang2021structure,ren2021combining} are proposed to estimate appearance flow between the references and desired targets. 
Models are trained with either unsupervised method or pre-calculated labels obtained by 3D models of human bodies.
Although the flow-based methods generate realistic details, they may fail to extract accurate motions when complex deformations or severe occlusions are observed.
Some other methods~\cite{zhang2020cross,Zhou_2021_CVPR} extract dense correspondences with the attention-based operation. 
They can generate accurate structures for the final images.
However, the quadratic memory footprint limits these methods to estimate high-resolution correspondence.
Our model with sparse attention can be applied to extract high-resolution neural textures without increasing the memory footprint dramatically.

\section{The Proposed Model}

In this paper, we propose a novel model for controllable person image synthesis. 
We introduce an efficient spatial transformation operation \emph{i.e.} neural texture extraction and distribution (NTED) operation in Sec.~\ref{sec:hook-tile}. 
In Sec.~\ref{sec:person_synthesis}, a generative model is designed with a hierarchical strategy that applies NTED operations at different scales. 
We introduce the loss functions in Sec.~\ref{sec:training_losses}.

\begin{figure*}
\centering
\includegraphics[width=1\linewidth]{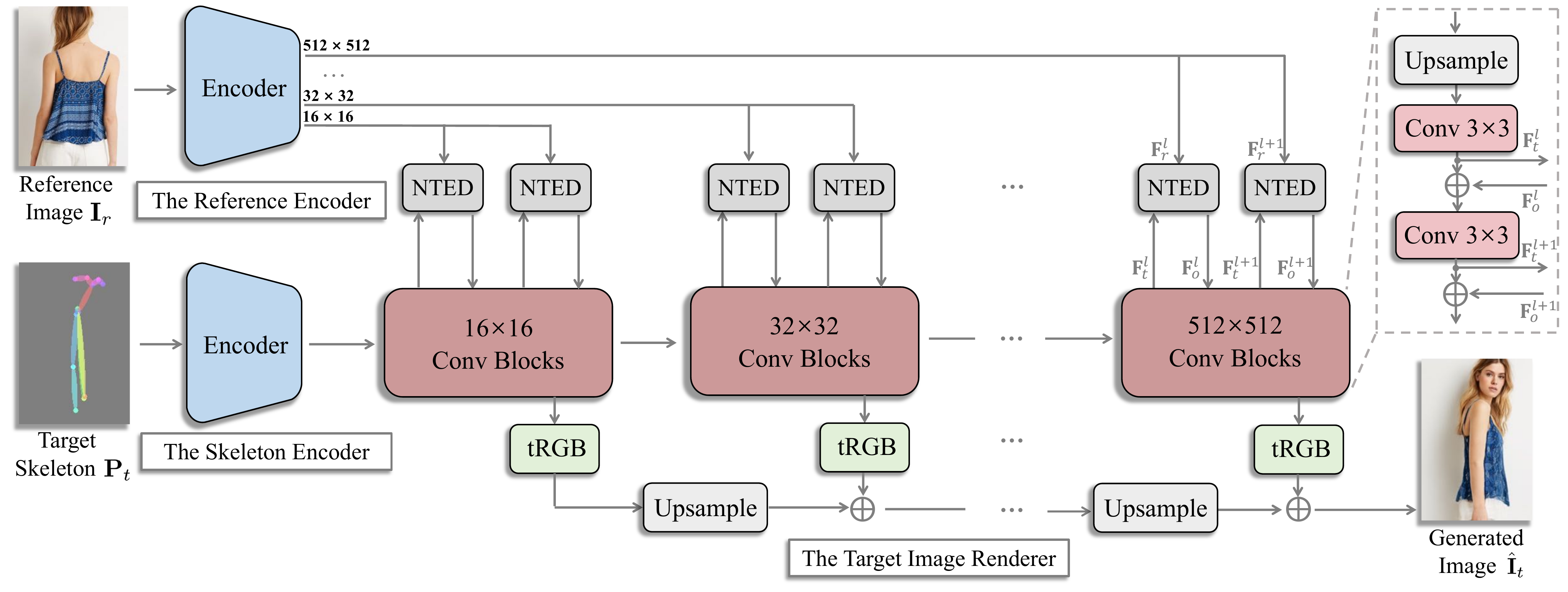}
\caption{Overview of the proposed model. Our model generates the result images by rendering target skeletons with reference features. NTED operations are used at different scales to deform both local and global contexts.}
 
\label{fig:network}
\end{figure*}

\subsection{The NTED Operation}
\label{sec:hook-tile}

A fundamental challenge of the person image synthesis task is to accurately reassemble the reference images.
In this subsection, we introduce a NTED operation.
As shown in Fig.~\ref{fig:hook-tile}, this operation consists of two steps: the extraction operation and the distribution operation.

\vspace{1mm}
\noindent
\textbf{The Extraction Operation} is responsible for extracting semantic neural textures from the reference feature maps. 
This operation is achieved by an attention step where each neural texture is calculated with a weighted sum of the values.
Let $\mathbf{F}_r \in \mathbb{R}^{hw \times c}$ represents the feature map extracted from the reference image $\mathbf{I}_r$.
Symbols $h$ and $w$ are the spatial sizes of the feature map.
The number of feature channels is denoted as $c$.
The attention correlation matrix is calculated between $\mathbf{F}_r$ and the semantic extraction filters $\mathbf{W}_{e} \in \mathbb{R}^{k \times c}$.
\begin{equation}
  \label{eq:extract}
  \mathbf{C}_{e} = \mathbf{W}_{e} \mathbf{F}_r^T
\end{equation}
where $\mathbf{C}_{e} \in \mathbb{R}^{k \times hw}$ is the correlation matrix. 
Each row $i$ of $\mathbf{C}_{e}$ contains the contributions of every reference feature to the $i^{th}$ neural texture.
The semantic extraction filters $\mathbf{W}_{e}$ are implemented using convolutional filters. The same filters are used for all images in a dataset.
This setting helps the model to automatically learn suitable semantic components. 
Meanwhile, the neural texture extracted by a specific filter always represents the same semantic component, which helps the model to disentangle the appearance of different semantics.




After obtaining $\mathbf{C}_e$, a softmax function is applied to normalize the correlation matrix across feature positions.
\begin{equation}
  \bar{\mathbf{C}}_{e}^{i, j} = \frac{\exp(\mathbf{C}_{e}^{i, j})}{\sum_{j=1}^{hw}\exp(\mathbf{C}_{e}^{i,j}) }
\end{equation}
where $\bar{\mathbf{C}}_{e}$ is the normalized correlation matrix. 
The neural textures are extracted by a weighted sum of the values.
\begin{equation}
  \mathbf{F}_e = \bar{\mathbf{C}}_{e} f(\mathbf{F}_r)
\end{equation}
where values $f(\mathbf{F}_r)$ is obtained by transforming $\mathbf{F}_r$ with a projection function $f$.
The neural textures $\mathbf{F}_e \in \mathbb{R}^{k \times c}$ represent the appearance of the semantic entities.




\vspace{1mm}
\noindent
\textbf{The Distribution Operation} is responsible for distributing the extracted neural textures according to the target poses.
Let $\mathbf{F}_t \in \mathbb{R}^{hw \times c}$ denotes the feature map of the target skeletons $\mathbf{P}_t$. 
The distribution operation first models the spatial distribution of the semantic neural textures.
\begin{equation}
  \mathbf{C}_d = \mathbf{W}_d \mathbf{F}_t^T
\end{equation}
where $\mathbf{W}_d \in \mathbb{R}^{k \times c}$ denotes the semantic distribution filters. Similar to that of the extraction operation, we implement $\mathbf{W}_d$ using convolutional filters. 
The output matrix $\mathbf{C}_d \in \mathbb{R}^{k \times hw}$ contains the correlations between all semantic entities and all target features. 
We normalize this matrix along with axis $k$.
\begin{equation}
\label{eq:norm_distribution}
  \bar{\mathbf{C}}_d^{i, j} = \frac{\exp(\mathbf{C}_{d}^{i, j})}{\sum_{i=1}^{k}\exp(\mathbf{C}_{d}^{i,j}) }
\end{equation}
Each column $j$ of $\bar{\mathbf{C}}_d$ represents the contributions of each semantic neural texture when generating $j^{th}$ features.
The final output of the NTED operation is calculated as 
\begin{equation}
  \mathbf{F}_o = \bar{\mathbf{C}}_d^T \mathbf{F}_e 
\end{equation}
where $\mathbf{F}_o \in \mathbb{R}^{hw \times c}$ is the output feature map.
To simplify the notation, we define a warping notation $\mathcal{W}$ to represent the overall NTED operation as 
\begin{equation}
\label{eq:final_nted}
  \mathbf{F}_o = \mathcal{W}(f(\mathbf{F}_r), \bar{\mathbf{C}}_{ed}) = \bar{\mathbf{C}}_d^T\bar{\mathbf{C}}_e f(\mathbf{F}_r)
\end{equation}
where $\bar{\mathbf{C}}_{ed} = \bar{\mathbf{C}}_d^T\bar{\mathbf{C}}_e$ denotes the deformations estimated by the NTED operation.
The NTED operation can be seen as a linear attention whose computational complexity is linear with the length of sequences. See \textit{Supplementary Materials} for more discussions.


\subsection{Person Image Synthesis Model}
\label{sec:person_synthesis}
We design the person image synthesis model as a pose-conditioned generative neural network that generates photo-realistic images $\hat{\mathbf{I}}_t$  by rendering the target skeletons $\mathbf{P}_t$ with the neural textures extracted from the reference images $\mathbf{I}_r$.  
The architecture is shown in Fig~\ref{fig:network}. It can be seen that this model is composed of three modules: the skeleton encoder, the reference encoder, and the target image renderer.

\vspace{1mm}
\noindent
\textbf{The Skeleton Encoder} 
is designed to transform the target skeletons into feature maps.
This encoder takes a skeleton representation with resolution $512 \times 512$. 
The final output of the encoder is with resolution $16 \times 16$.
A total of $5$ encoder blocks are contained in the encoder where each block down-samples the inputs with a factor of $2$.

\vspace{1mm}
\noindent
\textbf{The Reference Encoder} is responsible for encoding the reference images into multi-scale feature maps.
We use a similar architecture to the skeleton encoder. 
Feature maps are generated for each scale from $512 \times 512$ to $16 \times 16$.

\vspace{1mm}
\noindent
\textbf{The Target Image Renderer} is used to synthesize the target images by rendering the skeletons using the extracted neural textures.
This network takes the feature maps generated by the skeleton encoder as inputs. 
For each layer, the NTED operation is used to deform the reference features.
We design the NTED operation to predict the residual of current results.
The aligned feature map $\mathbf{F}_o^l$ of the $l^{th}$ NTED operation is added to the target feature map $\mathbf{F}_t^l$.
We employ the image skip connections proposed in StyleGAN2~\cite{karras2020analyzing}.
The RGB images are predicted at different scales. 
The final outputs are calculated by up-sampling and summing the contributions of these RGB outputs.



\subsection{Training Losses}
\label{sec:training_losses}
We train our model in an end-to-end manner to simultaneously learn the neural texture deformation and the target image generation.
We employ several loss functions that fulfill specific tasks.

\vspace{1mm}
\noindent
\textbf{Attention Reconstruction Loss $\mathcal{L}_{attn}$}. 
We use an attention reconstruction loss to constrain the NTED operation to extract accurate deformations.
This loss penalizes the $\ell_1$ difference between the deformed output and the ground truth image for each layer $l$.
\begin{equation}
  \mathcal{L}_{attn} = \sum_l \lVert \mathbf{I}_t^{l\downarrow} - \mathcal{W}(\mathbf{I}_r^{l\downarrow}, \bar{\mathbf{C}}_{ed}^l) \rVert_1
\end{equation}
where $\mathbf{I}_t^{l\downarrow}$ and $\mathbf{I}_r^{l\downarrow}$ are obtained by resizing the target images  $\mathbf{I}_t$ and the reference images $\mathbf{I}_r$ to the resolution of the $l^{th}$ layer. $\bar{\mathbf{C}}_{ed}^l$ represents the deformations estimated by the NTED operation in the $l^{th}$ layer.

\vspace{1mm}
\noindent
\textbf{Reconstruction Loss $\mathcal{L}_{rec}$}. 
A reconstruction loss is used to calculate the difference between the generated images $\hat{\mathbf{I}}_t$ and the ground-truth images $\mathbf{I}_t$. 
We employ the perceptual loss proposed in paper~\cite{johnson2016perceptual}.
\begin{equation}
\label{eq:rec}
  \mathcal{L}_{rec} = \sum_i \lVert \phi_i(\mathbf{I}_t) - \phi_i(\hat{\mathbf{I}}_t) \rVert_1 
\end{equation}
where $\phi_i$ denotes the $i$-th activation map of the pre-trained VGG-19 network.  
This loss calculates the $\ell_1$ difference between the VGG-19 activations.

\vspace{1mm}
\noindent
\textbf{Face Reconstruction Loss $\mathcal{L}_{face}$}. 
In addition to the reconstruction loss $\mathcal{L}_{rec}$, we also use a face reconstruction loss to calculate the perceptual distance between cropped faces.
\begin{equation}
\label{eq:face}
  \mathcal{L}_{face} = \sum_i \lVert \phi_i(C_{face}(\mathbf{I}_t)) - \phi_i(C_{face}(\hat{\mathbf{I}}_t)) \rVert_1 
\end{equation}
where $C_{face}$ is the face cropping function that crops the faces according to the target poses.

\vspace{1mm}
\noindent
\textbf{Adversarial Loss $\mathcal{L}_{adv}$}. A generative adversarial loss is employed to mimic the distribution of ground-truth images. 
A discriminator is trained to distinguish outputs from the real images in the target domain.
\begin{equation}
\label{eq:adv}
  \mathcal{L}_{adv} = \mathbb{E}[\log(1-D(G(\mathbf{P}_t, \mathbf{I}_r)))] + \mathbb{E}[\log(D(\mathbf{I}_t))]
\end{equation}
where $G$ and $D$ denote the generator and the discriminator.

\vspace{1mm}
\noindent
\textbf{Total Loss $\mathcal{L}_{total}$}. 
We train our model with a joint loss. 
\begin{equation}
  \mathcal{L}_{total} = \lambda_{attn} \mathcal{L}_{attn} + \lambda_{rec} \mathcal{L}_{rec} + \lambda_{face} \mathcal{L}_{face} + \mathcal{L}_{adv}
\end{equation}
where $\lambda_{attn}$, $\lambda_{rec}$, and $\lambda_{face}$ are the hyper-parameters.






\begin{figure}[t]
\begin{center}
\includegraphics[width=1\linewidth]{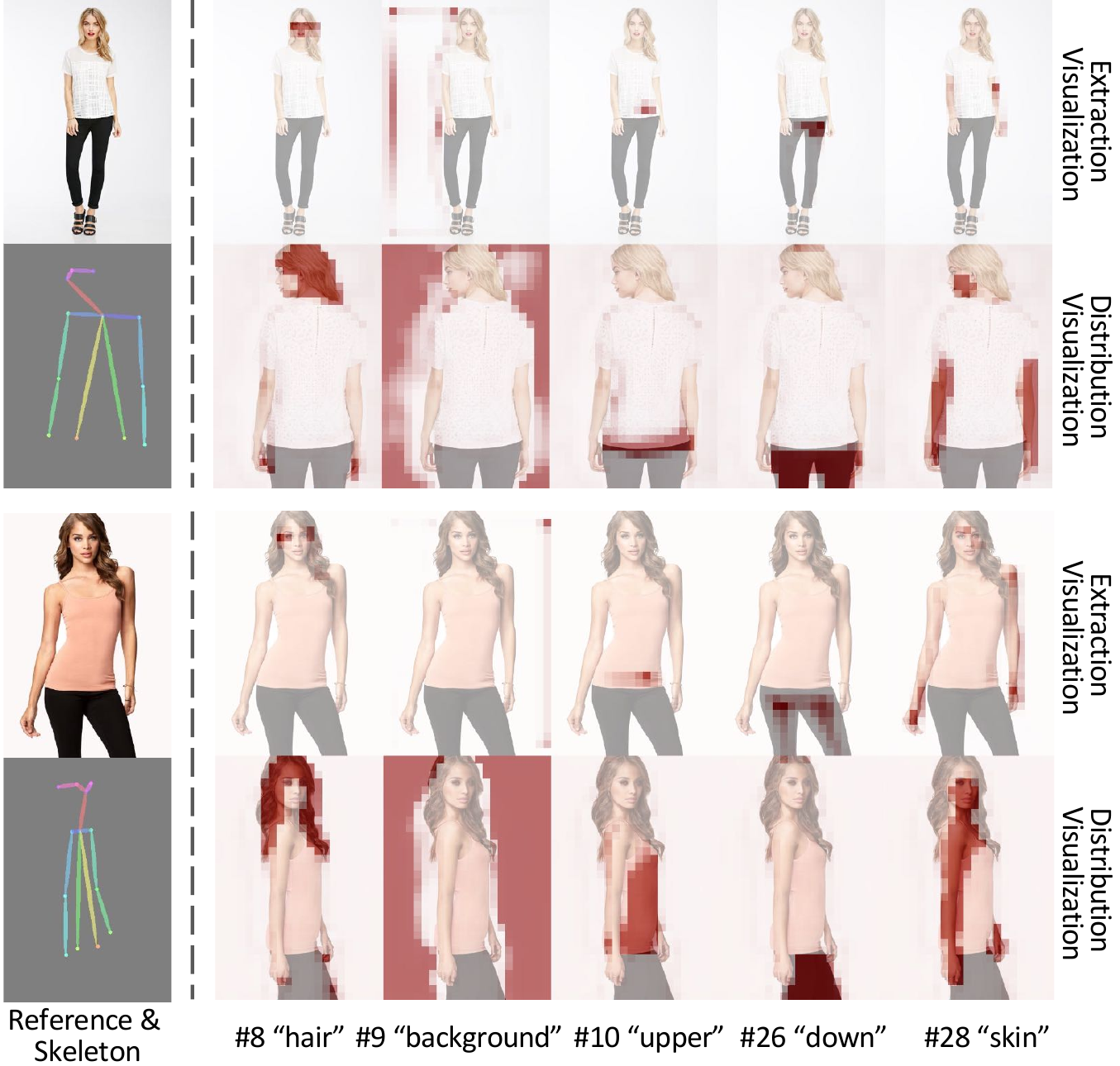}
\end{center}
   \caption{The visualizations of several typical channels in $\bar{\mathbf{C}}_e^l$ and $\bar{\mathbf{C}}_d^l$ at layer $l$ with resolution $64\times64$. For each sample, the first row is the visualizations of the extraction operation, while the second row is the visualizations of the distribution operation.}

\label{fig:visi}
\end{figure}

\begin{table*}[]
\setlength\extrarowheight{2pt}
\centering
\resizebox{.78\textwidth}{!}{%
\begin{tabular}{p{1.2cm}||P{1.6cm}P{1.6cm}P{1.6cm}P{1.6cm}P{1.6cm}||P{1.6cm}P{1.6cm}} \hline
      & \multicolumn{5}{c||}{$256\times176$ Images}                          & \multicolumn{2}{c}{$512\times352$ Images} \\ \cline{2-8} 
      & PATN     & ADGAN     & PISE     &  GFLA    & Ours            & CocosNet2       & Ours                      \\ \hline
SSIM $\uparrow$  & 0.6714    & 0.6735 & 0.6537 &  0.7082& \textbf{0.7182}  &   0.7236             & \textbf{0.7376}           \\
LPIPS $\downarrow$& 0.2533    & 0.2255 & 0.2244 &  0.1878& \textbf{0.1752} &   0.2265             & \textbf{0.1980}           \\
FID $\downarrow$  & 20.728    & 14.540 & 11.518 &  9.8272& \textbf{8.6838} &   13.325             & \textbf{7.7821}           \\ \hline
\end{tabular}
}

\caption{The quantitative comparisons with several state-of-the-art methods on both $256\times176$ and $512\times352$ images.}
\label{tab:object}
\end{table*}

\section{Optimization for Appearance Control}
\label{sec:appearance_control}

Given the trained model, images with arbitrary poses can be synthesized by extracting and reassembling neural textures of the reference images. 
Although we do not use any semantic labels to supervise the neural texture extraction, the proposed model can obtain meaningful and expressive latent vectors. 
Fig.~\ref{fig:visi} shows the visualizations of the attention correlation matrix $\bar{\mathbf{C}}_e$ and $\bar{\mathbf{C}}_d$.
It can be clearly seen that a specific neural texture is always formed by summing the regions with a certain semantic component and controls the generation of the corresponding target regions.
Therefore, we can expect to control the appearance of the final images by exchanging the corresponding semantic neural textures of different references.







Without loss of generality, we assume that a novel image $\hat{\mathbf{I}}_{t}$ is generated from two reference images $\mathbf{I}_{r1}$ and $\mathbf{I}_{r2}$ by using the semantic entity $i$ of $\mathbf{I}_{r2}$ and the other semantic components of $\mathbf{I}_{r1}$.
To achieve this goal, the neural textures related to the semantic entity $i$ are extracted from $\mathbf{I}_{r2}$, while the others are extracted from $\mathbf{I}_{r1}$.
Inspired by paper~\cite{lewis2021tryongan}, we use an optimization method to automatically implement this task.
Let $\mathbf{F}_{e1}^{[1, L]} \equiv \{\mathbf{F}_{e1}^1, \mathbf{F}_{e1}^2, ..., \mathbf{F}_{e1}^L\}$ and 
$\mathbf{F}_{e2}^{[1, L]} \equiv \{\mathbf{F}_{e2}^1, \mathbf{F}_{e2}^2, ..., \mathbf{F}_{e2}^L\}$ denote neural textures of $\mathbf{I}_{r1}$ and $\mathbf{I}_{r2}$.
Symbol $L$ is the number of network layers.
We define a set of mask tensor $\mathbf{m}^{[1, L]}\equiv \{\mathbf{m}^1, \mathbf{m}^2, ..., \mathbf{m}^L\}$ to interpolate between the extracted neural textures.
For each layer $l$, the fused neural textures are obtained by 
\begin{equation}
\label{eq:interp}
  \mathbf{F}_e^{l}=\mathbf{F}_{e1}^l+\mathbf{m}^l(\mathbf{F}_{e2}^l-\mathbf{F}_{e1}^l)
\end{equation}
where $\mathbf{m}^l \in \mathbb{R}^{k \times 1}$ has values between $0$ and $1$.
We optimize the interpolation coefficients $\mathbf{m}^{[1, L]}$ with
\begin{equation}
  \label{eq:opt}
  \mathcal{L}_{opt} = \lambda_{regu} \mathcal{L}_{regu} + \lambda_{r1} \mathcal{L}_{r1} + \lambda_{r2} \mathcal{L}_{r2}
\end{equation}

\noindent
\textbf{Regularization Loss $\mathcal{L}_{regu}$}. 
Desired coefficients $\mathbf{m}^{[1, L]}$ should be assigned with large values for the neural textures related to the semantic entity $i$ and small values for the other textures.
An operation $\mathcal{A}$ is defined to distinguish between the neural textures. 
Recalling that the attention correlation matrix $\bar{\mathbf{C}}_d \in \mathbb{R}^{k \times hw}$ of the distribution operation contains the spatial distributions of different semantic neural textures. 
It provides a clear clue to find neural textures generating semantic entity $i$. 
Let $\mathbf{S}_t$ denotes the binary segmentation labels of the generated images $\hat{\mathbf{I}}_t$ obtained by  off-the-shelf segmentation techniques, where the regions of the semantic entity $i$ are set as $1$. Operation $\mathcal{A}$ is defined as 
\begin{equation}
  \mathcal{A}(\bar{\mathbf{C}}_d, \mathbf{S}_t^{\downarrow})=\frac{\sum_{hw}\bar{\mathbf{C}}_d \odot \mathbf{S}_t^{\downarrow}}{\sum_{hw}\mathbf{S}_t^{\downarrow}} > \sigma
\end{equation}
where $\mathcal{A}(\bar{\mathbf{C}}_d, \mathbf{S}_t^{\downarrow}) \in \{0, 1 \}^{k\times 1}$ contains the indexes of the neural textures related to the semantic entity $i$. 
$\mathbf{S}_t^{\downarrow} \in \{0, 1 \}^{1 \times hw}$ is the resized segmentation labels.  
Symbol $\odot$ denotes the spatial-wise multiplication.
Operation $\mathcal{A}$ calculates the average attention coefficient in the regions of semantic entity $i$. 
The neural textures with attention values larger than a threshold $\sigma$ are regarded as the neural textures generating region $i$.
Our regularization loss is defined as

\begin{equation}
 \mathcal{L}_{regu} = \sum_l \mathcal{A} (\bar{\mathbf{C}}_d^{l}, \mathbf{S}_t^{l\downarrow})\odot(\mathbf{1}-\mathbf{m}^l) 
                             + \mathcal{A}(\bar{\mathbf{C}}_d^{l}, \mathbf{1}-\mathbf{S}_t^{l\downarrow})\odot\mathbf{m}^l
\end{equation}

\noindent
\textbf{Appearance Maintaining Loss $\mathcal{L}_{r1}$}. 
The appearance maintaining loss encourages the final image $\hat{\mathbf{I}}_t$ maintains the editing-irrelevant semantic components in $\mathbf{I}_{r1}$. 
Let $\hat{\mathbf{I}}_{t1}$ and $\mathbf{S}_{t1}$ denote the pose-transformed image of $\mathbf{I}_{r1}$ and its segmentation label. This loss calculates the perceptual distance between the masked $\hat{\mathbf{I}}_t$ and $\hat{\mathbf{I}}_{t1}$.
\begin{equation}
  \mathcal{L}_{r1} = \mathcal{L}_{rec}(\hat{\mathbf{I}}_t\odot(\mathbf{1}-\mathbf{S}_t), \hat{\mathbf{I}}_{t1}\odot(\mathbf{1}-\mathbf{S}_{t1}))
\end{equation}
where $\mathcal{L}_{rec}$ is the perceptual reconstruction loss in Eq.~\ref{eq:rec}.

\noindent
\textbf{Appearance Editing Loss $\mathcal{L}_{r2}$}. The appearance editing loss encourages the final image $\hat{\mathbf{I}}_t$ contains the semantic entity $i$ in $\mathbf{I}_{r2}$. Let $\hat{\mathbf{I}}_{t2}$ and $\mathbf{S}_{t2}$ denote the pose-transformed image of $\mathbf{I}_{r2}$ and its segmentation label. This loss calculates the perceptual distance between the masked $\hat{\mathbf{I}}_t$ and $\hat{\mathbf{I}}_{t2}$. 
\begin{equation}
  \mathcal{L}_{r2} = \mathcal{L}_{rec}(\hat{\mathbf{I}}_t\odot\mathbf{S}_t, \hat{\mathbf{I}}_{t2}\odot\mathbf{S}_{t2})
\end{equation}
With the joint loss function $\mathcal{L}_{opt}$ in Eq.~\ref{eq:opt}, we can optimize the interpolation coefficients $\mathbf{m}^{[1, L]}$. 
After obtaining $\mathbf{m}^{[1, L]}$, the fused neural textures $\mathbf{F}_e$ in Eq.~\ref{eq:interp} can be sent to the target image renderer to generate the editing results.

\section{Experiment}
In this section, experiments are conducted to evaluate the performance of the proposed model.
The implementation details are first provided in Sec.~\ref{sec:Implementation_Details}. 
Then, we compare our model with several state-of-the-art methods in Sec.~\ref{sec:comparisons}.
In Sec.~\ref{sec:Ablation_Study}, ablation models are trained to verify the efficacy of the proposed modules.
Finally, in Sec.~\ref{sec:appearance_control_results} we provide results of appearance control.

\subsection{Implementation Details}
\label{sec:Implementation_Details}
\noindent

\noindent
\textbf{Dataset}. We train our model on the In-shop Clothes Retrieval Benchmark of the DeepFashion dataset~\cite{liu2016deepfashion}.
This dataset contains $52,712$ high-resolution images of fashion models.
Images of the same person in the same cloth are paired for training and testing.  
The skeletons are extracted by OpenPose~\cite{8765346}. 
We use the dataset splits provided by~\cite{zhu2019progressive}. 
There are a total of $101,966$ pairs in the training set and $8,570$ pairs in the testing set. 
\noindent
\textbf{Metrics.} We evaluate the model performance from different aspects. \textit{Structure Similarity Index Measure} (SSIM)~\cite{wang2004image} and \textit{Learned Perceptual Image Patch Similarity} (LPIPS)~\cite{zhang2018unreasonable} are used to calculate the reconstruction accuracy. SSIM calculates the pixel-level image similarity, while LPIPS provides perceptual distance by employing a network trained on human judgments.
\textit{Fr\'echet Inception Distance} (FID)~\cite{heusel2017gans} is used to measure the realism of the generated images. It calculates the distance between the distributions of synthesized images and real images.

\noindent
\textbf{Training Details.} 
In our experiments, we train the proposed model with $256 \times 176$ and $512 \times 352$ images.
We use Adam~\cite{long2014convnets} solver with $\beta_1 = 0, \beta_2=0.99$. The learning rate is set to $2\times10^{-3}$ for both generator and discriminator. 
The model is trained for $200$ epochs with a batch size of $16$. 
More details can be found in the \textit{Supplementary Materials}.




\subsection{Comparisons}
\label{sec:comparisons}
We compare the proposed model with several state-of-the-art methods including
PATN\cite{zhu2019progressive}, ADGAN~\cite{men2020controllable}, GFLA~\cite{ren2020deep}, PISE~\cite{zhang2021pise}, and CocosNet2~\cite{Zhou_2021_CVPR}.
The released weights provided by the corresponding authors are used to obtain the results.

\noindent
\textbf{Quantitative Results.} 
The evaluation results are shown in Tab.~\ref{tab:object}. 
We evaluate the performance on both $256\times176$ images and $512\times352$ images according to the training set of the competitors. 
Since CoCosv2 uses a different train/test split, we retrain this model using their source codes.
It can be seen that our model achieves the best results compared with the state-of-the-art methods.
This means that our model can generate images with not only accurate structures but also realistic details.

\begin{figure}[t]
\centering
\includegraphics[width=1\linewidth]{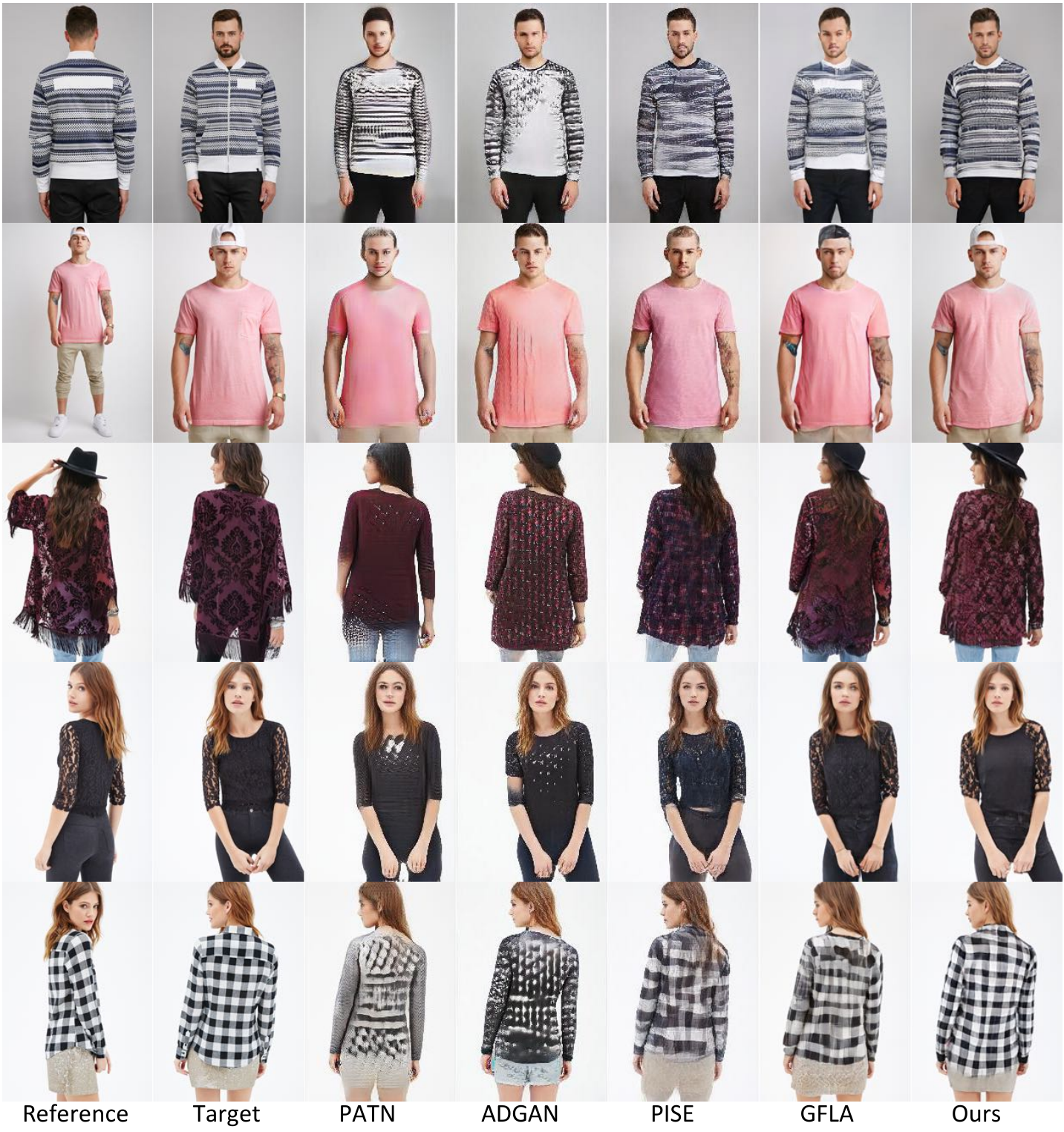}
\caption{Qualitative comparisons with several state-of-the-art methods on the DeepFashion dataset with $256\times176$ images.}
 
\label{fig:comparison_256}
\end{figure}

\begin{figure}[t]
\centering
\includegraphics[width=0.93\linewidth]{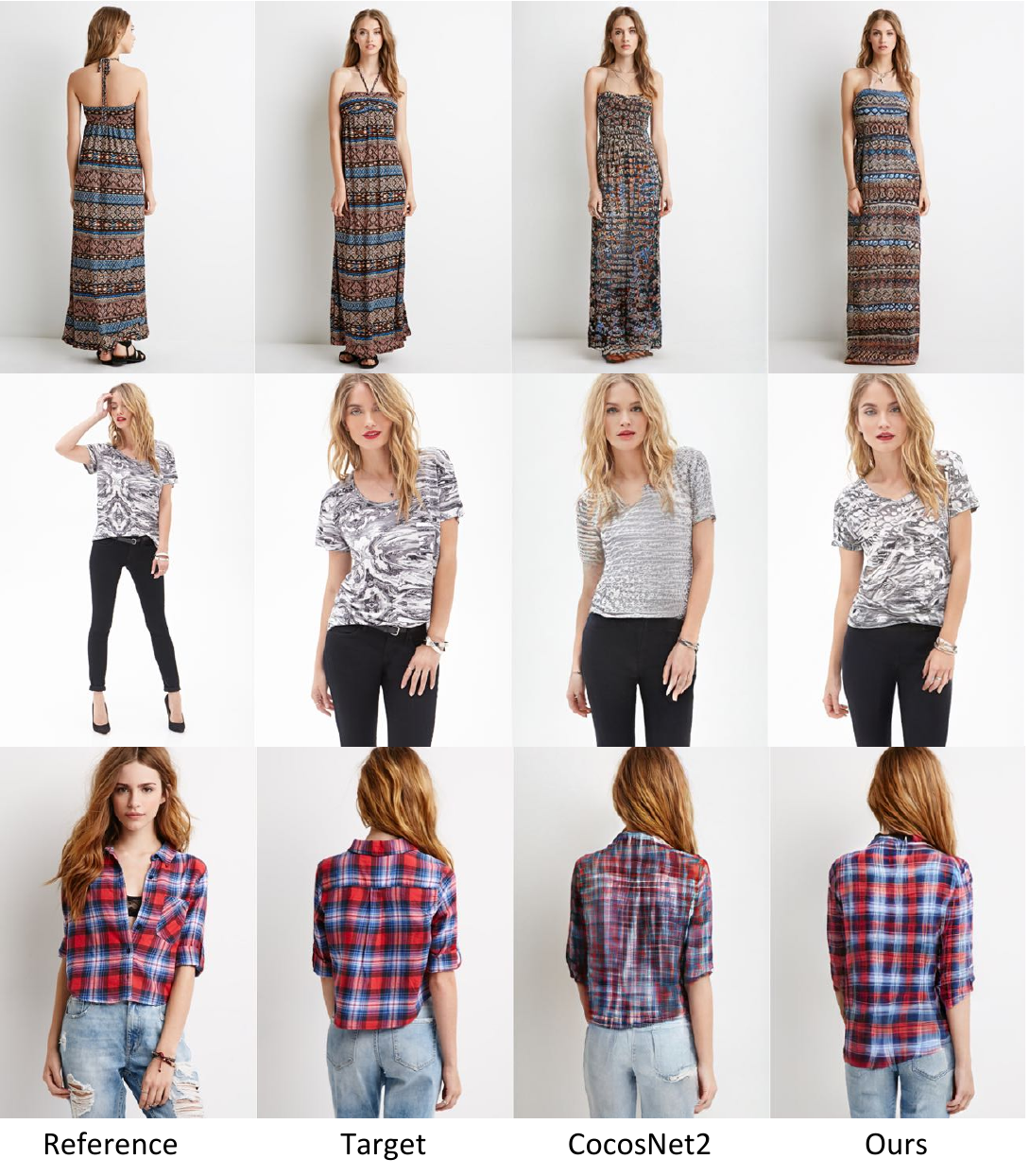}
 \caption{Qualitative comparisons with CocosNet2 on the DeepFashion dataset with $512\times352$ images.}
 
\label{fig:comparison_512}
\end{figure}

\noindent
\textbf{Qualitative Results.} 
We provide the generated results in Fig.~\ref{fig:comparison_256} and Fig.~\ref{fig:comparison_512}.
It can be seen that PATN struggles to generate realistic images due to the lack of efficient spatial deformation blocks.
PATN and ADGAN generate images with accurate structures. However, they extract image appearance using 1D vectors, which hinders the generation of complex textures.
The flow-based method GFLA can generate realistic textures. However, it yields noticeable artifacts when severe occlusions are observed.
CocosNet2 generates high-resolution images with accurate structures. 
However, it fails to maintain the patterns of complex textures. 
Our model generates visually appealing results with both accurate structures and vivid textures.

\subsection{Ablation Study}
\label{sec:Ablation_Study}
We evaluate the efficacy of the proposed neural texture extraction and distribution operation by comparing our model with several variants.

\noindent
\textbf{Baseline Model.}
A baseline model is trained to prove the necessity of the neural texture deformation module.
An auto-encoder network is used for this model.
The reference images and target skeletons are concatenated as the model inputs.
We train this model using the reconstruction loss, the face reconstruction loss, and the adversarial loss.

\noindent
\textbf{Style-based Model.}
A style-based model is designed to compare the NTED operation with the style-based modulation block proposed in StyleGAN2.
In this model, the NTED operations are replaced by the style modulation blocks. 
Reference images are encoded as 1D vectors to modulate the generation.
We train this model using the same loss functions as that of the Baseline Model.

\begin{figure}[t]
\centering
\includegraphics[width=0.95\linewidth]{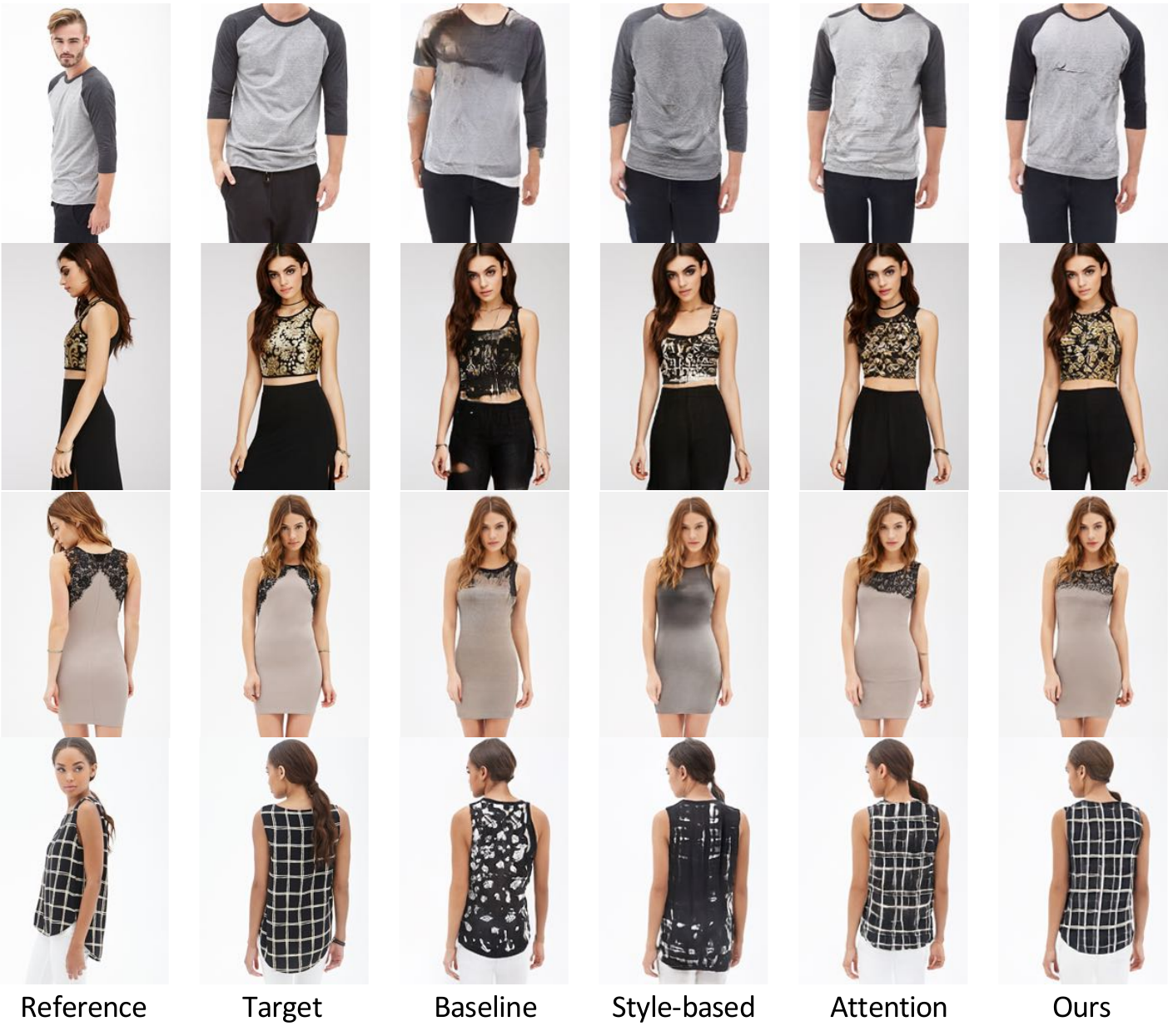}
 \caption{Qualitative results of the ablation study.}
\vspace{-3mm}
\label{fig:ablation}
\end{figure}

\noindent
\textbf{Attention Model.} The attention model is used to compare the NTED operation with the vanilla attention operation. 
We replace our NTED operations with the attention operations. 
The attention correlations are calculated between the reference feature $\mathbf{F}_r$ and the target skeleton feature $\mathbf{F}_t$.
To ensure the fairness of the comparison, we do not use the sub-sampling trick. 
Meanwhile, the number of feature channels is not reduced when calculating the attention.
The model is trained with the same loss functions as our method.

\noindent
\textbf{Ours.} We employ the proposed model with the NTED operations here.

We train all ablation models with the same setting as that of our model.
The quantitative results of the ablation study are shown in Tab.~\ref{tab:ablation}.
It can be seen that our model achieves competitive results compared with the ablation methods.
Taking the advantage of the generative adversarial techniques, the baseline model generates realistic person images with a good FID score.
However, the poor LPIPS result indicates that the model cannot faithfully reconstruct the textures due to the lack of efficient spatial transformation blocks. 
The style-based model improves the LPIPS score by leveraging both local and global contexts.
However, the 1D vectors are insufficient to represent complex spatial distributions, which may lead to performance degradation.
The attention model tries to establish the correlations between all sources to all targets. 
However, as discussed above, each target position only needs to sample a local source patch,
which implies that some calculations may be unnecessary.
This inference can be confirmed by comparing the evaluation results of the attention model with ours.
Our model achieves competitive results with less than half FLOPs of the attention model.

We show the qualitative results in Fig.~\ref{fig:ablation}. 
It can be seen that the Baseline Model fails to reproduce complex spatial distributions.
The style-based model alleviates this problem by hierarchically injecting the extracted vectors.
However, the uniform modulation hinders it to generate local details.
The attention model and our model can faithfully reconstruct the textures of reference images.

\begin{table}[]
\setlength\extrarowheight{1pt}
\centering
\resizebox{0.5\textwidth}{!}{%
\begin{tabular}{p{1.3cm}||P{1.8cm}P{1.8cm}P{1.8cm}P{1.8cm}}
\hline 
                   & Baseline & Style-based & Attention       & Ours \\ \hline
                   &          &            &                  &                     \\ [-11pt] \hline
SSIM $\uparrow$    & 0.7085   & 0.7111     & 0.7158           & \textbf{0.7182}     \\
LPIPS $\downarrow$ & 0.1935   & 0.1884     & 0.1761           & \textbf{0.1752}     \\
FID $\downarrow$   & 8.6568   & 9.3502     & \textbf{8.5732}  & 8.6838      \\ 
FLOPs $\downarrow$ & \textbf{53.73 G}   & 62.57 G     & 219.94 G    & 103.99 G      \\ \hline
\end{tabular}%
}
\caption{The evaluation results of the ablation study.}
\label{tab:ablation}
\end{table}

\begin{figure}[t]
\centering
\includegraphics[width=1\linewidth]{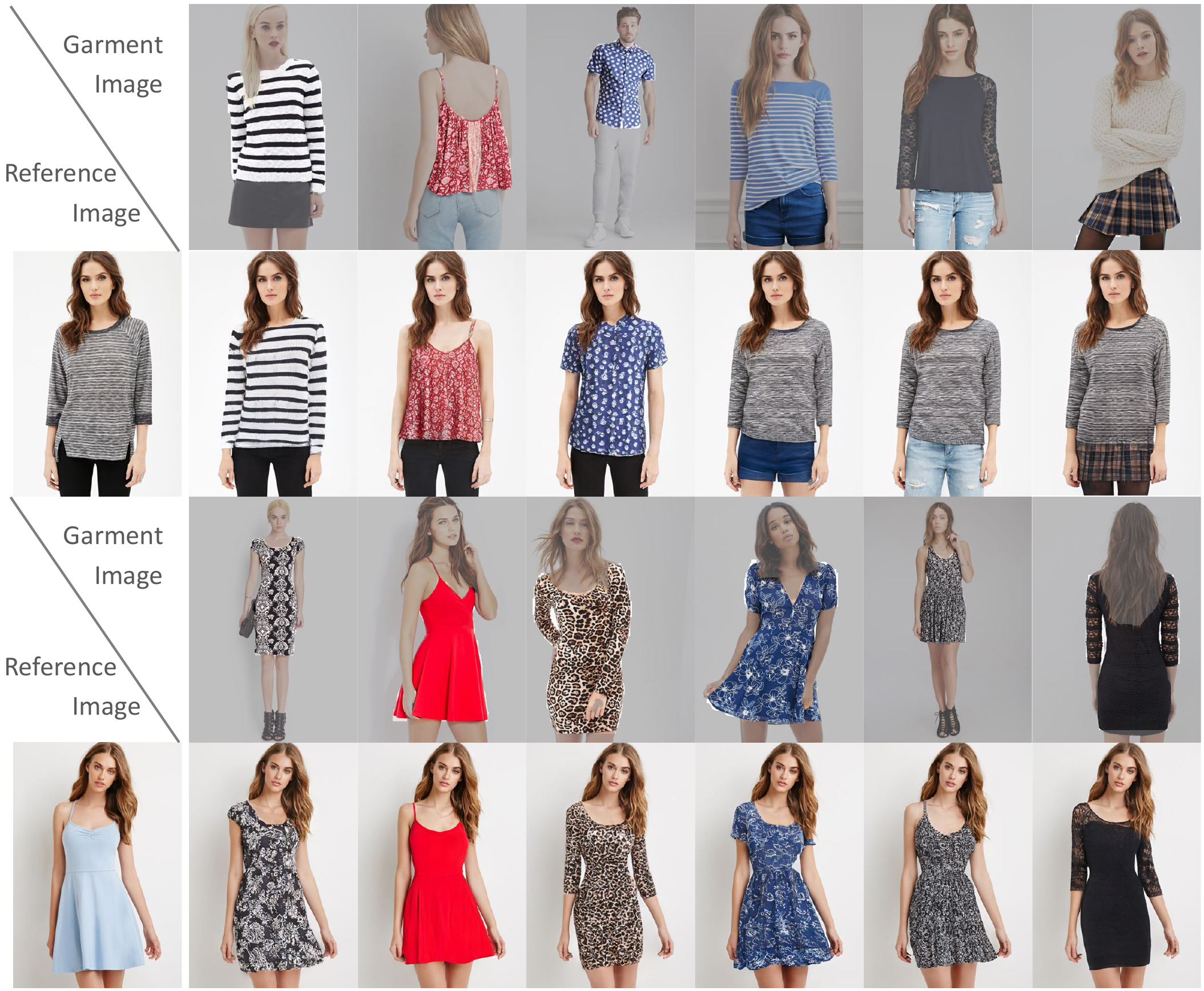}
\caption{Images generated by controlling the appearance of interested areas. For each sample, the first row contains the garment images. The second row contains the generated images.}
\label{fig:appearance}
\end{figure}

\subsection{Appearance Control Results}
\label{sec:appearance_control_results}
Our model enables appearance control by combining the neural textures extracted from different reference images.
We optimize the interpolation coefficients by using the methods described in Sec~\ref{sec:appearance_control}.
The results are shown in Fig.~\ref{fig:appearance}. We observe that our model can seamlessly combine the areas of interest and generate coherent images.
The garments are extracted from images with arbitrary poses. 
Both structure and textures are faithfully reconstructed. 
Meanwhile, the unrelated semantic regions are well-preserved, which indicates that our model represents different semantics with disentangled neural textures.

\begin{figure}[t]
\centering
\includegraphics[width=1\linewidth]{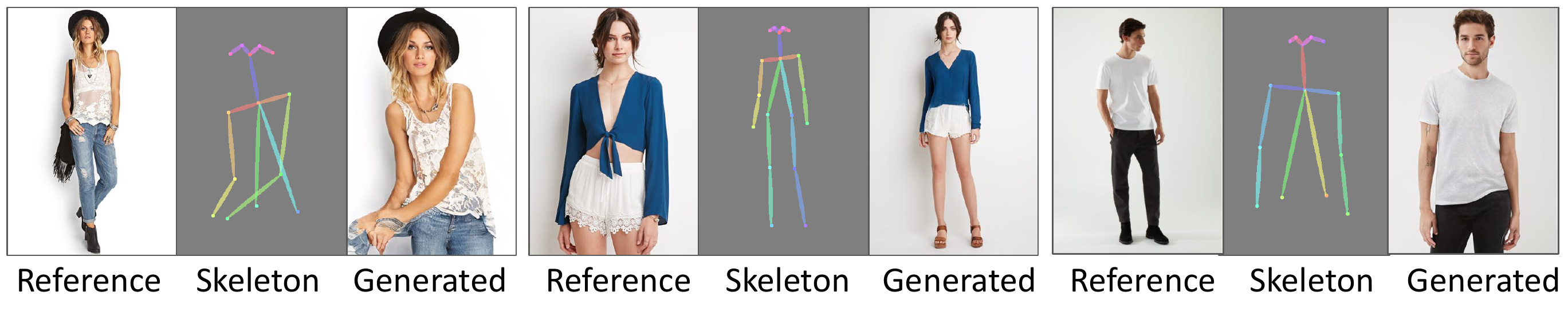}
\caption{Failure cases caused by underrepresented poses (left), garments (middle), and in-the-wild identities (right).}
\label{fig:failure}
\end{figure}

\section{Conclusion and Discussion}
We have presented a novel model for synthesizing photo-realistic person images by explicitly controlling the pose and appearance of a reference image.  
The hook-tile operation is described for neural texture deformation. 
This operation first extracts hierarchical semantic neural textures from reference images. 
Then the extracted neural textures are reassembled according to the spatial distributions learned from the target poses. 
Our model outperforms state-of-the-art methods and generates high-resolution realistic images even for references with extremely complex textures. 
Meanwhile, the disentangled neural textures enable a further application on appearance control.
Promising results are generated by seamlessly merging the areas of interest from different images.

\noindent
\textbf{Limitations.} Although our model generates promising results, it still fails in cases of underrepresented images. 
We show some failure cases in Fig.~\ref{fig:failure}. Artifacts or inconsistencies can be found in these results. 
We believe that training the model on diverse data will alleviate this problem. 
Meanwhile, flexibly extracting local patches by modeling the correlations of the adjacent deformations will improve the model generalization, which will be our further work.

\noindent
\textbf{Ethical Considerations.} 
The pose transfer or appearance control applications could be misused and pose a societal threat. We do not condone using our work with the intent of spreading misinformation or tarnishing reputation.

\vspace{4mm}
\noindent
\textbf{Acknowledgment.}
This work was supported by 
National Natural Science Foundation of China (No. 62172021) and Shenzhen Fundamental Research Program (GXWD20201231165807007-20200806163656003) 

{\small
\bibliographystyle{ieee_fullname}
\bibliography{egbib}

\begin{thebibliography}{10}\itemsep=-1pt

\bibitem{albahar2021pose}
Badour AlBahar, Jingwan Lu, Jimei Yang, Zhixin Shu, Eli Shechtman, and Jia-Bin
  Huang.
\newblock Pose with style: Detail-preserving pose-guided image synthesis with
  conditional stylegan.
\newblock {\em arXiv preprint arXiv:2109.06166}, 2021.

\bibitem{8765346}
Z. {Cao}, G. {Hidalgo Martinez}, T. {Simon}, S. {Wei}, and Y.~A. {Sheikh}.
\newblock Openpose: Realtime multi-person 2d pose estimation using part
  affinity fields.
\newblock {\em IEEE Transactions on Pattern Analysis and Machine Intelligence},
  2019.

\bibitem{chen20182}
Yunpeng Chen, Yannis Kalantidis, Jianshu Li, Shuicheng Yan, and Jiashi Feng.
\newblock Double attention networks.
\newblock {\em arXiv preprint arXiv:1810.11579}, 2018.

\bibitem{choi2018stargan}
Yunjey Choi, Minje Choi, Munyoung Kim, Jung-Woo Ha, Sunghun Kim, and Jaegul
  Choo.
\newblock Stargan: Unified generative adversarial networks for multi-domain
  image-to-image translation.
\newblock In {\em Proceedings of the IEEE conference on computer vision and
  pattern recognition}, pages 8789--8797, 2018.

\bibitem{esser2018variational}
Patrick Esser, Ekaterina Sutter, and Bj{\"o}rn Ommer.
\newblock A variational u-net for conditional appearance and shape generation.
\newblock In {\em Proceedings of the IEEE Conference on Computer Vision and
  Pattern Recognition}, pages 8857--8866, 2018.

\bibitem{goodfellow2016deep}
Ian Goodfellow, Yoshua Bengio, Aaron Courville, and Yoshua Bengio.
\newblock {\em Deep learning}, volume~1.
\newblock MIT press Cambridge, 2016.

\bibitem{heusel2017gans}
Martin Heusel, Hubert Ramsauer, Thomas Unterthiner, Bernhard Nessler, and Sepp
  Hochreiter.
\newblock Gans trained by a two time-scale update rule converge to a local nash
  equilibrium.
\newblock In {\em Advances in Neural Information Processing Systems}, pages
  6626--6637, 2017.

\bibitem{huang2018multimodal}
Xun Huang, Ming-Yu Liu, Serge Belongie, and Jan Kautz.
\newblock Multimodal unsupervised image-to-image translation.
\newblock In {\em Proceedings of the European conference on computer vision
  (ECCV)}, pages 172--189, 2018.

\bibitem{isola2017image}
Phillip Isola, Jun-Yan Zhu, Tinghui Zhou, and Alexei~A Efros.
\newblock Image-to-image translation with conditional adversarial networks.
\newblock In {\em Proceedings of the IEEE conference on computer vision and
  pattern recognition}, pages 1125--1134, 2017.

\bibitem{johnson2016perceptual}
Justin Johnson, Alexandre Alahi, and Li Fei-Fei.
\newblock Perceptual losses for real-time style transfer and super-resolution.
\newblock In {\em European conference on computer vision}, pages 694--711.
  Springer, 2016.

\bibitem{karras2020analyzing}
Tero Karras, Samuli Laine, Miika Aittala, Janne Hellsten, Jaakko Lehtinen, and
  Timo Aila.
\newblock Analyzing and improving the image quality of stylegan.
\newblock In {\em Proceedings of the IEEE/CVF Conference on Computer Vision and
  Pattern Recognition}, pages 8110--8119, 2020.

\bibitem{lewis2021tryongan}
Kathleen~M Lewis, Srivatsan Varadharajan, and Ira Kemelmacher-Shlizerman.
\newblock Tryongan: body-aware try-on via layered interpolation.
\newblock {\em ACM Transactions on Graphics (TOG)}, 40(4):1--10, 2021.

\bibitem{li2019dense}
Yining Li, Chen Huang, and Chen~Change Loy.
\newblock Dense intrinsic appearance flow for human pose transfer.
\newblock In {\em Proceedings of the IEEE/CVF Conference on Computer Vision and
  Pattern Recognition}, pages 3693--3702, 2019.

\bibitem{liu2019liquid}
Wen Liu, Zhixin Piao, Jie Min, Wenhan Luo, Lin Ma, and Shenghua Gao.
\newblock Liquid warping gan: A unified framework for human motion imitation,
  appearance transfer and novel view synthesis.
\newblock In {\em Proceedings of the IEEE/CVF International Conference on
  Computer Vision}, pages 5904--5913, 2019.

\bibitem{liu2016deepfashion}
Ziwei Liu, Ping Luo, Shi Qiu, Xiaogang Wang, and Xiaoou Tang.
\newblock Deepfashion: Powering robust clothes recognition and retrieval with
  rich annotations.
\newblock In {\em Proceedings of the IEEE conference on computer vision and
  pattern recognition}, pages 1096--1104, 2016.

\bibitem{long2014convnets}
Jonathan~L Long, Ning Zhang, and Trevor Darrell.
\newblock Do convnets learn correspondence?
\newblock {\em Advances in neural information processing systems},
  27:1601--1609, 2014.

\bibitem{ma2018disentangled}
Liqian Ma, Qianru Sun, Stamatios Georgoulis, Luc Van~Gool, Bernt Schiele, and
  Mario Fritz.
\newblock Disentangled person image generation.
\newblock In {\em Proceedings of the IEEE Conference on Computer Vision and
  Pattern Recognition}, pages 99--108, 2018.

\bibitem{men2020controllable}
Yifang Men, Yiming Mao, Yuning Jiang, Wei-Ying Ma, and Zhouhui Lian.
\newblock Controllable person image synthesis with attribute-decomposed gan.
\newblock In {\em Proceedings of the IEEE/CVF Conference on Computer Vision and
  Pattern Recognition}, pages 5084--5093, 2020.

\bibitem{mirza2014conditional}
Mehdi Mirza and Simon Osindero.
\newblock Conditional generative adversarial nets.
\newblock {\em arXiv preprint arXiv:1411.1784}, 2014.

\bibitem{park2019semantic}
Taesung Park, Ming-Yu Liu, Ting-Chun Wang, and Jun-Yan Zhu.
\newblock Semantic image synthesis with spatially-adaptive normalization.
\newblock In {\em Proceedings of the IEEE/CVF Conference on Computer Vision and
  Pattern Recognition}, pages 2337--2346, 2019.

\bibitem{ren2021combining}
Yurui Ren, Yubo Wu, Thomas~H Li, Shan Liu, and Ge Li.
\newblock Combining attention with flow for person image synthesis.
\newblock In {\em Proceedings of the 29th ACM International Conference on
  Multimedia}, pages 3737--3745, 2021.

\bibitem{ren2020deep}
Yurui Ren, Xiaoming Yu, Junming Chen, Thomas~H Li, and Ge Li.
\newblock Deep image spatial transformation for person image generation.
\newblock In {\em Proceedings of the IEEE/CVF Conference on Computer Vision and
  Pattern Recognition}, pages 7690--7699, 2020.

\bibitem{salimans2016improved}
Tim Salimans, Ian Goodfellow, Wojciech Zaremba, Vicki Cheung, Alec Radford, and
  Xi Chen.
\newblock Improved techniques for training gans.
\newblock {\em arXiv preprint arXiv:1606.03498}, 2016.

\bibitem{sarkar2021style}
Kripasindhu Sarkar, Vladislav Golyanik, Lingjie Liu, and Christian Theobalt.
\newblock Style and pose control for image synthesis of humans from a single
  monocular view.
\newblock {\em arXiv preprint arXiv:2102.11263}, 2021.

\bibitem{shen2021efficient}
Zhuoran Shen, Mingyuan Zhang, Haiyu Zhao, Shuai Yi, and Hongsheng Li.
\newblock Efficient attention: Attention with linear complexities.
\newblock In {\em Proceedings of the IEEE/CVF Winter Conference on Applications
  of Computer Vision}, pages 3531--3539, 2021.

\bibitem{siarohin2018deformable}
Aliaksandr Siarohin, Enver Sangineto, St{\'e}phane Lathuiliere, and Nicu Sebe.
\newblock Deformable gans for pose-based human image generation.
\newblock In {\em Proceedings of the IEEE Conference on Computer Vision and
  Pattern Recognition}, pages 3408--3416, 2018.

\bibitem{tang2021structure}
Jilin Tang, Yi Yuan, Tianjia Shao, Yong Liu, Mengmeng Wang, and Kun Zhou.
\newblock Structure-aware person image generation with pose decomposition and
  semantic correlation.
\newblock {\em arXiv preprint arXiv:2102.02972}, 2021.

\bibitem{vaswani2017attention}
Ashish Vaswani, Noam Shazeer, Niki Parmar, Jakob Uszkoreit, Llion Jones,
  Aidan~N Gomez, Lukasz Kaiser, and Illia Polosukhin.
\newblock Attention is all you need.
\newblock {\em arXiv preprint arXiv:1706.03762}, 2017.

\bibitem{wang2019few}
Ting-Chun Wang, Ming-Yu Liu, Andrew Tao, Guilin Liu, Jan Kautz, and Bryan
  Catanzaro.
\newblock Few-shot video-to-video synthesis.
\newblock {\em arXiv preprint arXiv:1910.12713}, 2019.

\bibitem{wang2018high}
Ting-Chun Wang, Ming-Yu Liu, Jun-Yan Zhu, Andrew Tao, Jan Kautz, and Bryan
  Catanzaro.
\newblock High-resolution image synthesis and semantic manipulation with
  conditional gans.
\newblock In {\em Proceedings of the IEEE conference on computer vision and
  pattern recognition}, pages 8798--8807, 2018.

\bibitem{wang2018non}
Xiaolong Wang, Ross Girshick, Abhinav Gupta, and Kaiming He.
\newblock Non-local neural networks.
\newblock In {\em Proceedings of the IEEE conference on computer vision and
  pattern recognition}, pages 7794--7803, 2018.

\bibitem{wang2004image}
Zhou Wang, Alan~C Bovik, Hamid~R Sheikh, and Eero~P Simoncelli.
\newblock Image quality assessment: from error visibility to structural
  similarity.
\newblock {\em IEEE transactions on image processing}, 13(4):600--612, 2004.

\bibitem{yu2019multi}
Xiaoming Yu, Yuanqi Chen, Shan Liu, Thomas Li, and Ge Li.
\newblock Multi-mapping image-to-image translation via learning
  disentanglement.
\newblock In {\em Advances in Neural Information Processing Systems}, 2019.

\bibitem{zhang2019self}
Han Zhang, Ian Goodfellow, Dimitris Metaxas, and Augustus Odena.
\newblock Self-attention generative adversarial networks.
\newblock In {\em International conference on machine learning}, pages
  7354--7363. PMLR, 2019.

\bibitem{zhang2021pise}
Jinsong Zhang, Kun Li, Yu-Kun Lai, and Jingyu Yang.
\newblock Pise: Person image synthesis and editing with decoupled gan.
\newblock {\em arXiv preprint arXiv:2103.04023}, 2021.

\bibitem{zhang2020cross}
Pan Zhang, Bo Zhang, Dong Chen, Lu Yuan, and Fang Wen.
\newblock Cross-domain correspondence learning for exemplar-based image
  translation.
\newblock In {\em Proceedings of the IEEE/CVF Conference on Computer Vision and
  Pattern Recognition}, pages 5143--5153, 2020.

\bibitem{zhang2018unreasonable}
Richard Zhang, Phillip Isola, Alexei~A Efros, Eli Shechtman, and Oliver Wang.
\newblock The unreasonable effectiveness of deep features as a perceptual
  metric.
\newblock In {\em Proceedings of the IEEE Conference on Computer Vision and
  Pattern Recognition}, pages 586--595, 2018.

\bibitem{Zhou_2021_CVPR}
Xingran Zhou, Bo Zhang, Ting Zhang, Pan Zhang, Jianmin Bao, Dong Chen, Zhongfei
  Zhang, and Fang Wen.
\newblock Cocosnet v2: Full-resolution correspondence learning for image
  translation.
\newblock In {\em Proceedings of the IEEE/CVF Conference on Computer Vision and
  Pattern Recognition (CVPR)}, pages 11465--11475, 2021.

\bibitem{zhu2017unpaired}
Jun-Yan Zhu, Taesung Park, Phillip Isola, and Alexei~A Efros.
\newblock Unpaired image-to-image translation using cycle-consistent
  adversarial networks.
\newblock In {\em Proceedings of the IEEE international conference on computer
  vision}, pages 2223--2232, 2017.

\bibitem{zhu2017multimodal}
Jun-Yan Zhu, Richard Zhang, Deepak Pathak, Trevor Darrell, Alexei~A Efros,
  Oliver Wang, and Eli Shechtman.
\newblock Multimodal image-to-image translation by enforcing bi-cycle
  consistency.
\newblock In {\em Advances in neural information processing systems}, pages
  465--476, 2017.

\bibitem{zhu2019progressive}
Zhen Zhu, Tengteng Huang, Baoguang Shi, Miao Yu, Bofei Wang, and Xiang Bai.
\newblock Progressive pose attention transfer for person image generation.
\newblock In {\em Proceedings of the IEEE/CVF Conference on Computer Vision and
  Pattern Recognition}, pages 2347--2356, 2019.

\end{thebibliography}
}

\clearpage
\onecolumn
\begin{alphasection}

\title{ Neural Texture Extraction and Distribution for  Controllable Person Image Synthesis \\ \textit{Supplementary Material}}

\author{Yurui Ren$^{1}$~~~~~Xiaoqing Fan$^{1}$~~~~~{Ge Li \footnotesize{\Letter}}$^{1}$~~~~~Shan Liu$^{2}$~~~~Thomas H. Li$^{3,1}$\\
$^1$School of Electronic and Computer Engineering, Peking University~~~$^2$Tencent America~~~\\
$^3$Advanced Institute of Information Technology, Peking University~~\\
{\tt\small ~~~yrren@pku.edu.cn~~~fanxiaoqing@stu.pku.edu.cn~~~geli@ece.pku.edu.cn}\\
{\tt\small shanl@tencent.com~~~tli@aiit.org.cn~~~}}

\maketitle

\section{Implementation Details}

\subsection{Network Architecture} 
The detailed architecture of the proposed model is shown in Fig.~\ref{fig:network_detail_encoder} and Fig.~\ref{fig:network_detail_decoder}. 
We take the model generating $512 \times 512$ images as an example to show the details.
The Neural Texture Extraction and Distribution (NTED) operations are used to deform multi-scale reference features.
We extract and distribute features at $16 \times 16$, $32 \times 32$, $64 \times 64$, $128 \times 128$, and $256 \times 256$ for generating $512 \times 512$ images.

\noindent
\textbf{The Skeleton Encoder.} 
The architecture of the skeleton encoder is shown in Fig.~\ref{fig:network_detail_encoder}. 
This network is used to extract target features $\mathbf{F}_t$ from the target skeletons $\mathbf{P}_t$.
We use the 2D heatmap created from the extracted human keypoints as the representation of the target skeletons.
Each heatmap channel corresponds to a particular part of the skeletons.
A basic encoder block containing a convolutional layer with stride $2$ is designed for this network.
A total of $5$ encoder blocks are contained in the encoder where each block down-samples the inputs with a factor of $2$.
The shape of the final output $\mathbf{F}_t$ is $16 \times 16 \times 512$.

\noindent
\textbf{The Reference Encoder.} 
The architecture of the reference encoder is shown in Fig.~\ref{fig:network_detail_encoder}. 
We design the reference encoder using a similar architecture to the skeleton encoder. 
Multi-scale features $\mathbf{F}_r^l$ are extracted from the reference images $\mathbf{I}_r$.
We record the reference features generated by every encoder block.
A total of $5$ feature maps representing both global and local contexts are obtained.

\noindent
\textbf{The Target Image Renderer.}
The target image renderer is shown in Fig.~\ref{fig:network_detail_decoder}. 
Target features $\mathbf{F}_t$ representing the desired poses are rendered as realistic images using the neural textures extracted from the reference features.
Instead of using $1 \times 1$ convolutional filters in the NTED operations, 
we allow to use 2D convolutional filters with spatial sizes to extract accurate semantics.
We define the number of semantics of different scales as $16$, $32$, $32$, $64$, and $64$. 
An explanation for the process of the target image renderer is provided.
For each layer, the network predicts the semantic distributions of the target images according to $\mathbf{F}_t^l$.
The predicted distributions are used to reassemble the extracted semantic neural textures and generate the deformed reference feature maps $\mathbf{F}_o^l$.
Target feature maps $\mathbf{F}_t^{l+1}$ with more details about the reference individuals are generated. 
These feature maps further help the network to predict more accurate spatial distributions and generate more realistic images in the next layer.

\begin{figure*}[ht]
\renewcommand\thefigure{\Alph{alphasect}.\arabic{figure}}
\begin{center}
\includegraphics[width=0.67\linewidth]{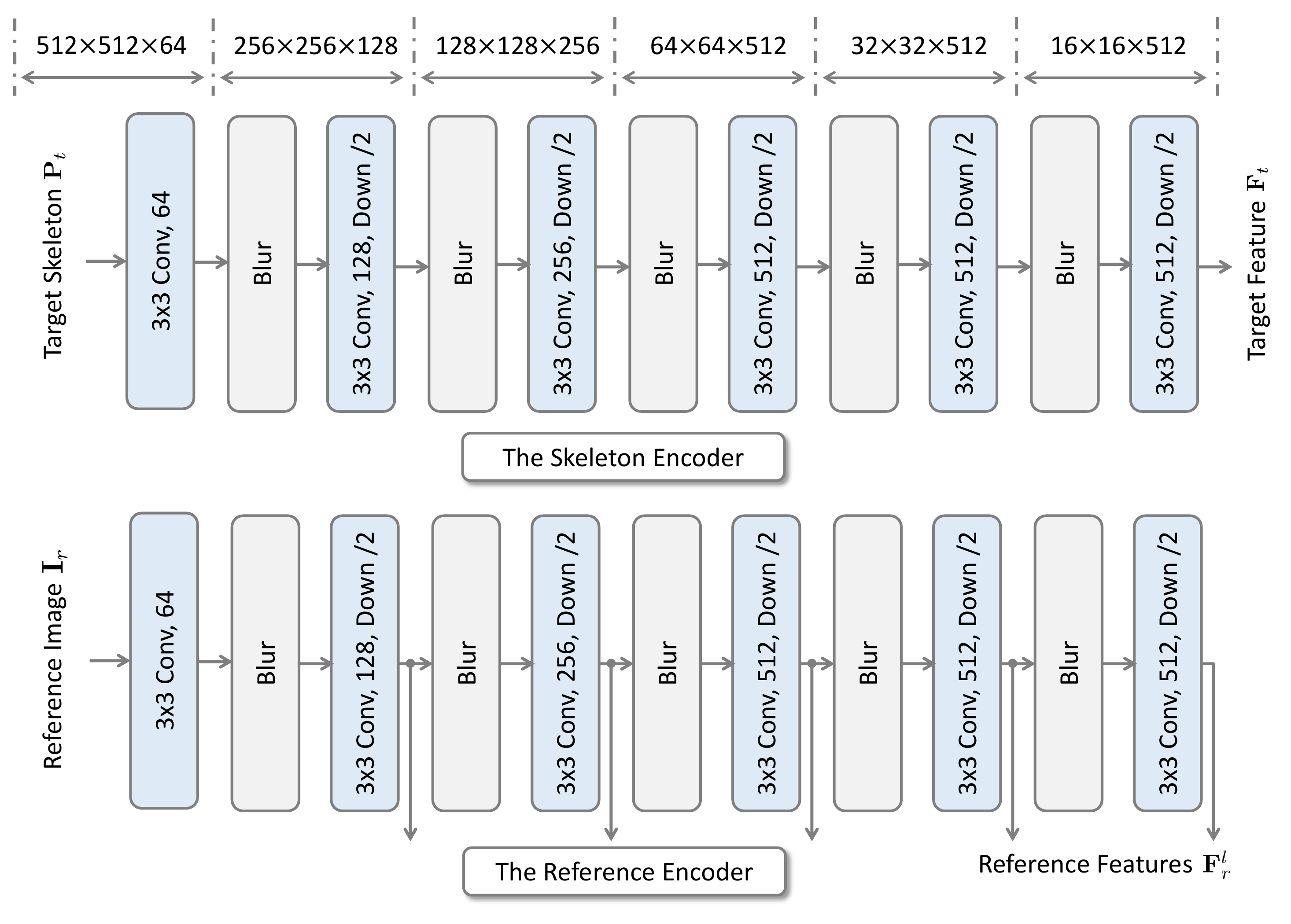}
\end{center}
   \caption{Detailed architecture of the skeleton encoder and the reference encoder. We mark the shape ($h\times w\times c$) of the corresponding features at the top of the figure.}

\label{fig:network_detail_encoder}
\end{figure*}

\subsection{Training Details} 
We train the model in an end-to-end manner to simultaneously learn the neural texture deformation and the target image generation.
We use the historical average technique~\cite{salimans2016improved} to update the average model by weighted averaging current parameters with previous parameters. 
The model is trained for $200$ epochs with a batch size of $16$. 
For the first $20k$ iterations, we do not use the adversarial loss for the training stability.
We use $\lambda_{rec} = 2$, $\lambda_{face} = 1$, $\lambda_{attn} = 15$, and $\lambda_{adv} = 1.5$ when training $256 \times 176$ images.
For training $512 \times 352$ images, we calculate perceptual loss at a number of different resolutions by applying pyramid downsampling on $\mathbf{I}_t$ and $\hat{\mathbf{I}}_t$. Images with resolutions of $512 \times 352$, $256 \times 176$, and $128 \times 88$ are used for this loss. The parameter of the reconstruction loss is $\lambda_{rec} = 1$.

\begin{figure*}[ht]
\renewcommand\thefigure{\Alph{alphasect}.\arabic{figure}}
\begin{center}
\includegraphics[width=1\linewidth]{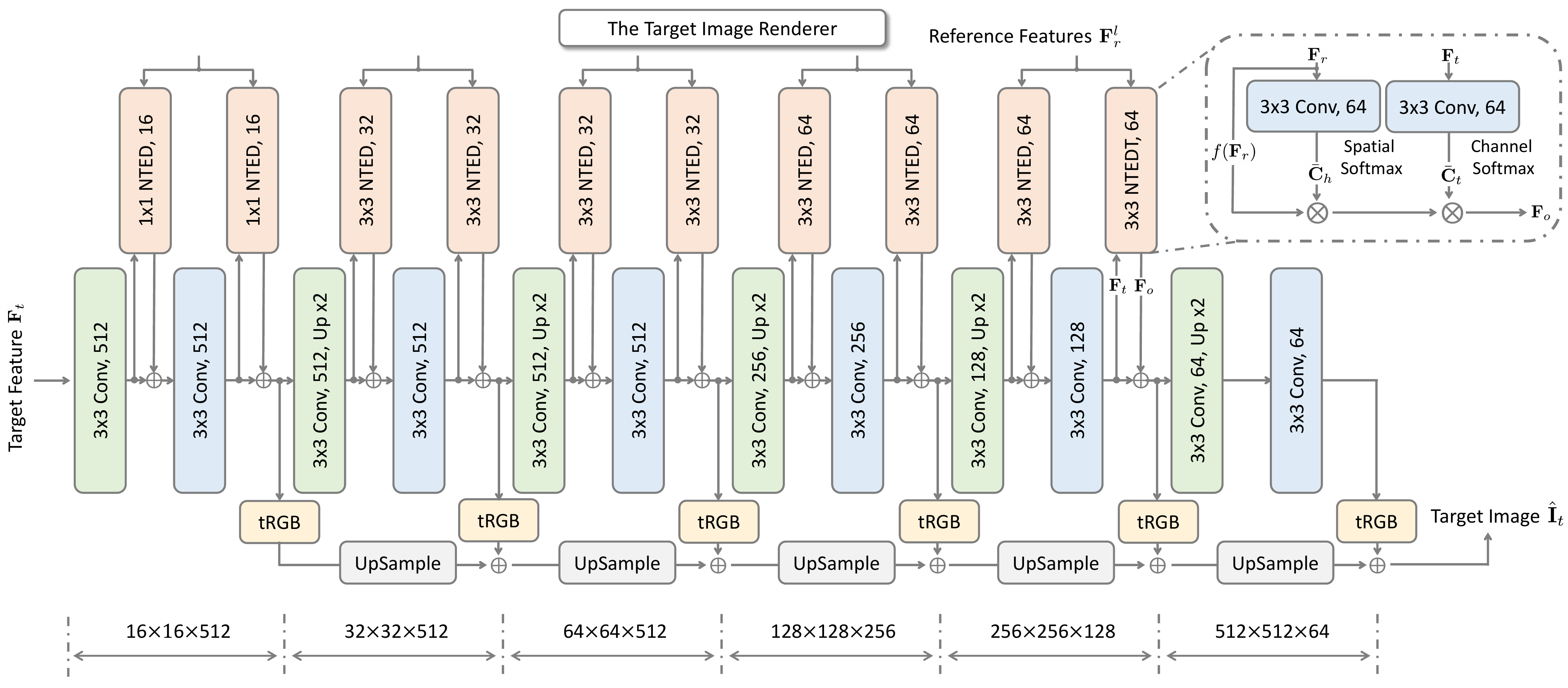}
\end{center}
   \caption{Detailed architecture of the target image renderer. We mark the shape ($h\times w\times c$) of the corresponding features at the bottom of the figure.}
 
\label{fig:network_detail_decoder}
\end{figure*}

\clearpage
\subsection{Details of Appearance Control}
\label{sec:appearance_control}

After training the model, image appearance can be explicitly controlled by fusing the appearance of different reference images.
In this section, we provide more explanations about the proposed appearance control method. 
This method is shown in Fig.~\ref{fig:appearance_control}.  
Semantic neural textures are first extracted from $\mathbf{I}_{r1}$ and $\mathbf{I}_{r2}$. 
Then, neural textures related to the interested garments are selected from $\mathbf{F}_{e2}$ to replace the corresponding neural textures in $\mathbf{F}_{e1}$.
We propose an optimization method to automatically search for the interpolation coefficients.
All model parameters are fixed in this optimization process. 
The interpolation coefficients are optimized according to the joint loss function $\mathcal{L}_{opt}$. 
We use three loss functions for different purposes.
The regularization loss $\mathcal{L}_{regu}$ is used to encourage the coefficients to assign large values for the neural textures related to the interested semantics and small values for the others. The appearance maintaining loss $\mathcal{L}_{r1}$ is used to maintain the editing-irrelevant semantic components in the reference images $\mathbf{I}_{r1}$. The appearance editing loss $\mathcal{L}_{r2}$ encourages the final images contain the corresponding appearance of the interested areas in $\mathbf{I}_{r2}$.
After obtaining the interpolation coefficients, we can calculate the fused neural textures $\mathbf{F}_e$ and generate the final images $\hat{\mathbf{I}}_t$ in arbitrary poses.
We use hyperparameter $\lambda_{regu}=1$, $\lambda_{r1}=3\times10^5$, and $\lambda_{r2}=9\times10^5$. 
Meaningful results can be obtained with less than $200$ iterations.

\begin{figure*}[ht]
\renewcommand\thefigure{\Alph{alphasect}.\arabic{figure}}
\centering
\includegraphics[width=0.8\linewidth]{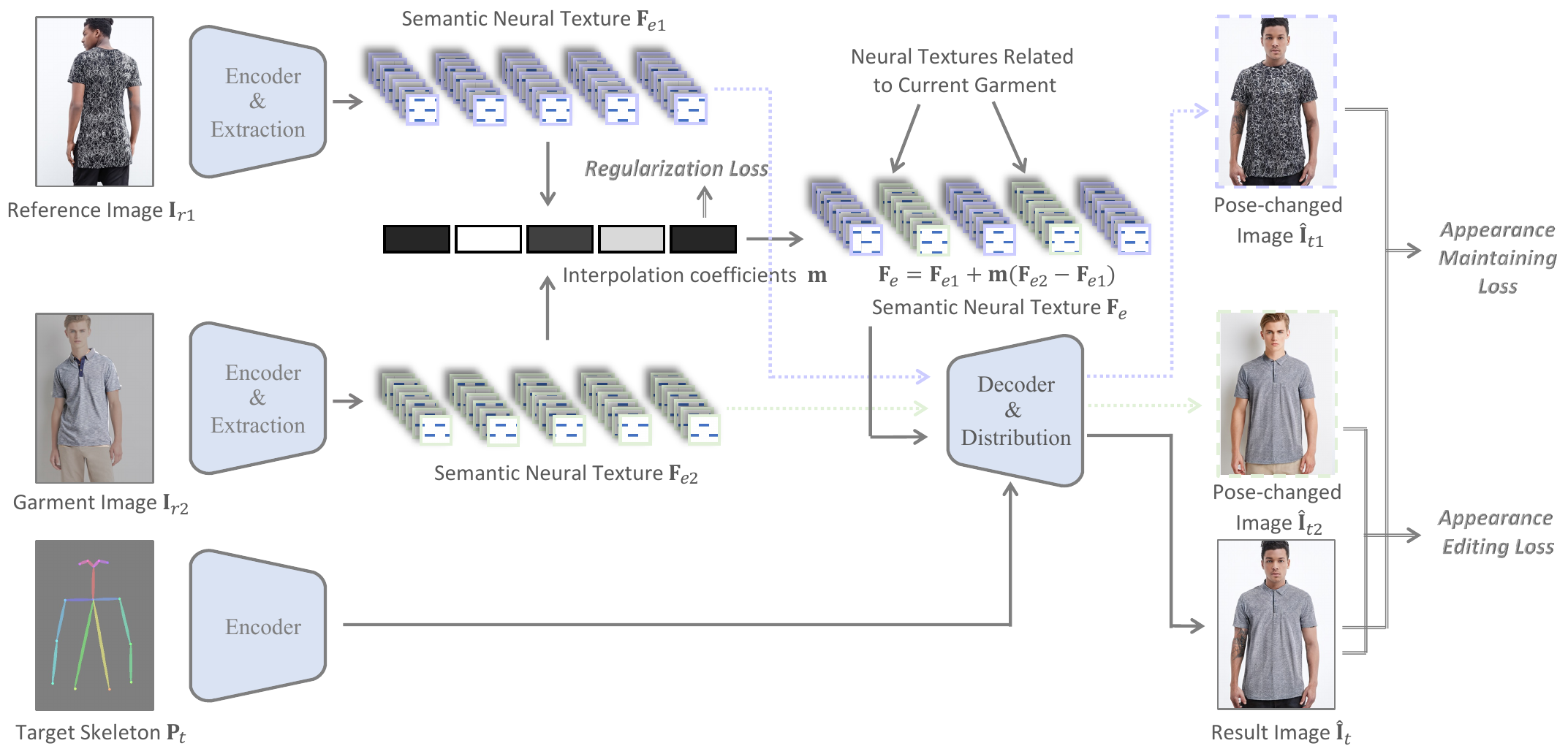}
\caption{Optimization method for appearance control. Model parameters are fixed. The interpolation coefficients are optimized according to the loss functions. The result images can be obtained by fusing the extracted neural textures $\mathbf{F}_{e1}$ and $\mathbf{F}_{e2}$ with the optimized interpolation coefficients.}

\label{fig:appearance_control}
\end{figure*}

\section{Complexity Analysis of the NTED Operation}

In this section, we analysis the complexity of the proposed NTED operation and compare it with the vanilla attention operation. 
Recall that this task requires to build correlations between the reference features $\mathbf{F}_r \in \mathbb{R}^{hw\times c}$ and the target features $\mathbf{F}_t \in \mathbb{R}^{hw\times c}$. 
The vanilla attention calculates outputs $\mathbf{F}_o$ by building dense correlations.
\begin{equation}
\renewcommand\theequation{\Alph{alphasect}.\arabic{equation}}
    \mathbf{F}_o=\mathcal{S}_j(\mathbf{F}_t\mathbf{F}_r^T)f(\mathbf{F}_r)
\end{equation}
where $\mathcal{S}_j$ represents a softmax function normalizing inputs along columns.
Due to the nonlinear function $\mathcal{S}_j$, this operation must first calculates $\mathbf{F}_t\mathbf{F}_r^T$ with $\mathcal{O}((wh)^2)$ computation complexity. Therefore, the complexity will increase dramatically, when the sequence length increases. 
The proposed NTED operation can be seen as a linear attention. 
According to Eq.~\ref{eq:final_nted}, it calculates the outputs as 
\begin{equation}
\renewcommand\theequation{\Alph{alphasect}.\arabic{equation}}
  \mathbf{F}_o = \bar{\mathbf{C}}_d^T\bar{\mathbf{C}}_e f(\mathbf{F}_r)=\mathcal{S}_i(\mathbf{W}_d\mathbf{F}_t^T)^T\mathcal{S}_j(\mathbf{W}_e\mathbf{F}_r^T)f(\mathbf{F}_r)
\end{equation}
In this operation, we can first calculate the neural textures with $\mathbf{F}_e = \bar{\mathbf{C}}_ef(\mathbf{F}_r)$ and then obtain the final outputs as $\mathbf{F}_o = \bar{\mathbf{C}}_d^T\mathbf{F}_e$. This operation avoid building direct interactions between $\mathbf{F}_t$ and $\mathbf{F}_r$.
We compare the complexity with the vanilla attention operation in Tab.~\ref{tab:complexity}.
It can be seen that our model has $\mathcal{O}(hwk+hwc)$ memory complexity and $\mathcal{O}(hwkc+hwc^2)$ computation complexity. Because $k$ is much smaller than $hw$, the NTED operation can significantly reduce the resource usage and enable efficient neural texture deformation.

\begin{table*}[ht]
\setlength\extrarowheight{2pt}
\centering
\renewcommand\thetable{\Alph{alphasect}.\arabic{table}}
\begin{tabular}
{P{4cm}P{3cm}P{3cm}}\\ \Xhline{2\arrayrulewidth}
                  & NTED Operation         & Attention Operation \\\hline
Memory complexity & $\mathcal{O}(hwk+hwc)$   & $\mathcal{O}(hwc+(hw)^2)$         \\
Comp. complexity  & $\mathcal{O}(hwkc+hwc^2)$   & $\mathcal{O}(hwc^2+c(hw)^2)$         \\ \Xhline{2\arrayrulewidth}
\end{tabular}
\caption{Comparisons of resource usage. Symbols $hw$ denote the spatial size of feature maps, $c$ is the number of channels, and $k$ is the number of semantics. In our setting, we define $hw \gg k$.}
\label{tab:complexity}
\end{table*}


\section{Analysis of the Extracted Neural Textures}
\label{analysis}
Our model is able to extract semantically meaningful neural textures from the reference images. 
In this section, we verify this by showing the neural textures used to generate specific semantics.
The normalized attention correlations $\bar{\mathbf{C}}_t \in \mathbb{R}^{k\times hw}$ obtained in Eq.~\ref{eq:norm_distribution} contains the spatial distribution of the neural textures.
Each column $j$ of $\bar{\mathbf{C}}_t$ contains the contributions of every neural textures when generating the $j^{th}$ feature. Therefore, we average the columns of a region $\mathbf{S} \in \mathbb{R}^{1 \times hw}$ to calculate $\bar{\mathbf{c}}_t$.
\begin{equation}
\renewcommand\theequation{\Alph{alphasect}.\arabic{equation}}
  \bar{\mathbf{c}}_t = \frac{\sum_{hw}\bar{\mathbf{C}}_t \odot \mathbf{S}^{\downarrow}}{\sum_{hw}\mathbf{S}^{\downarrow}}
\end{equation}
where $\bar{\mathbf{c}}_t \in \mathbb{R}^{k\times 1}$ indicates what neural textures are used to generate region $\mathbf{S}$.
We show $\bar{\mathbf{c}}_t$ of different regions in Fig.~\ref{fig:neural_texture}. It can be seen that a specific semantic entity is always generated by specific neural textures. 
Meanwhile, different semantic regions are expressed by different neural textures. 



\begin{figure*}[ht]
\renewcommand\thefigure{\Alph{alphasect}.\arabic{figure}}
\begin{center}
\includegraphics[width=0.6\linewidth]{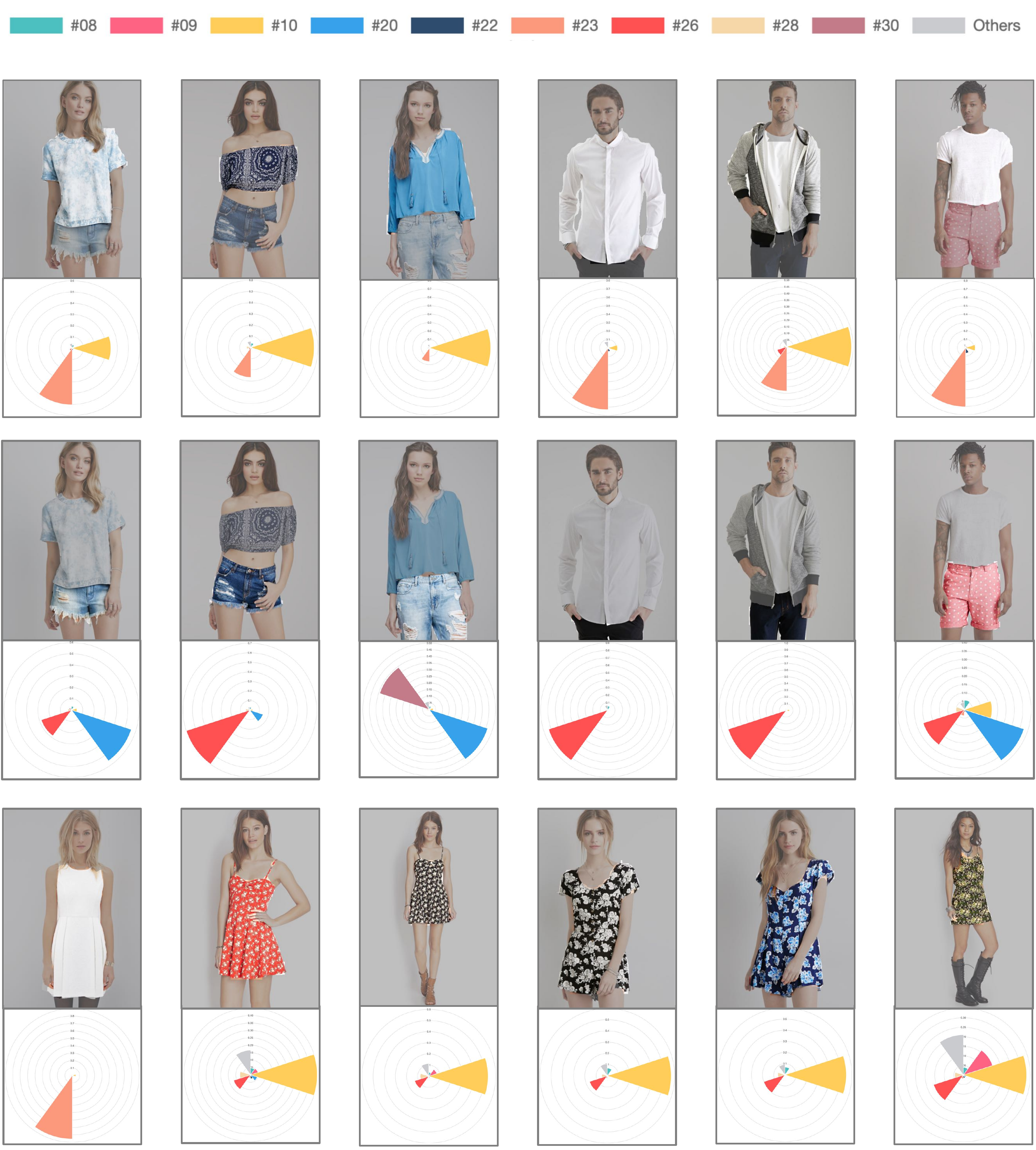}
\end{center}
   \caption{Neural textures related to specific semantic entities. Large color region indicates high coefficients. From up to down: upper clothes, pants, and dresses.}

\label{fig:neural_texture}
\end{figure*}


\newpage

\section{Additional Results} 
\label{sec:additional_results}

\subsection{Additional Results of Pose Control} 
We provide additional results of pose control in Fig.~\ref{fig:pose_control_0}, Fig.~\ref{fig:pose_control_1}, Fig.~\ref{fig:pose_control_2}, and Fig.~\ref{fig:pose_control_3}. 
We generate person images at the resolution $512 \times 352$ on the DeepFashion dataset~\cite{liu2016deepfashion}.

\begin{figure*}[ht]
\renewcommand\thefigure{\Alph{alphasect}.\arabic{figure}}
\begin{center}
\includegraphics[width=0.75\linewidth]{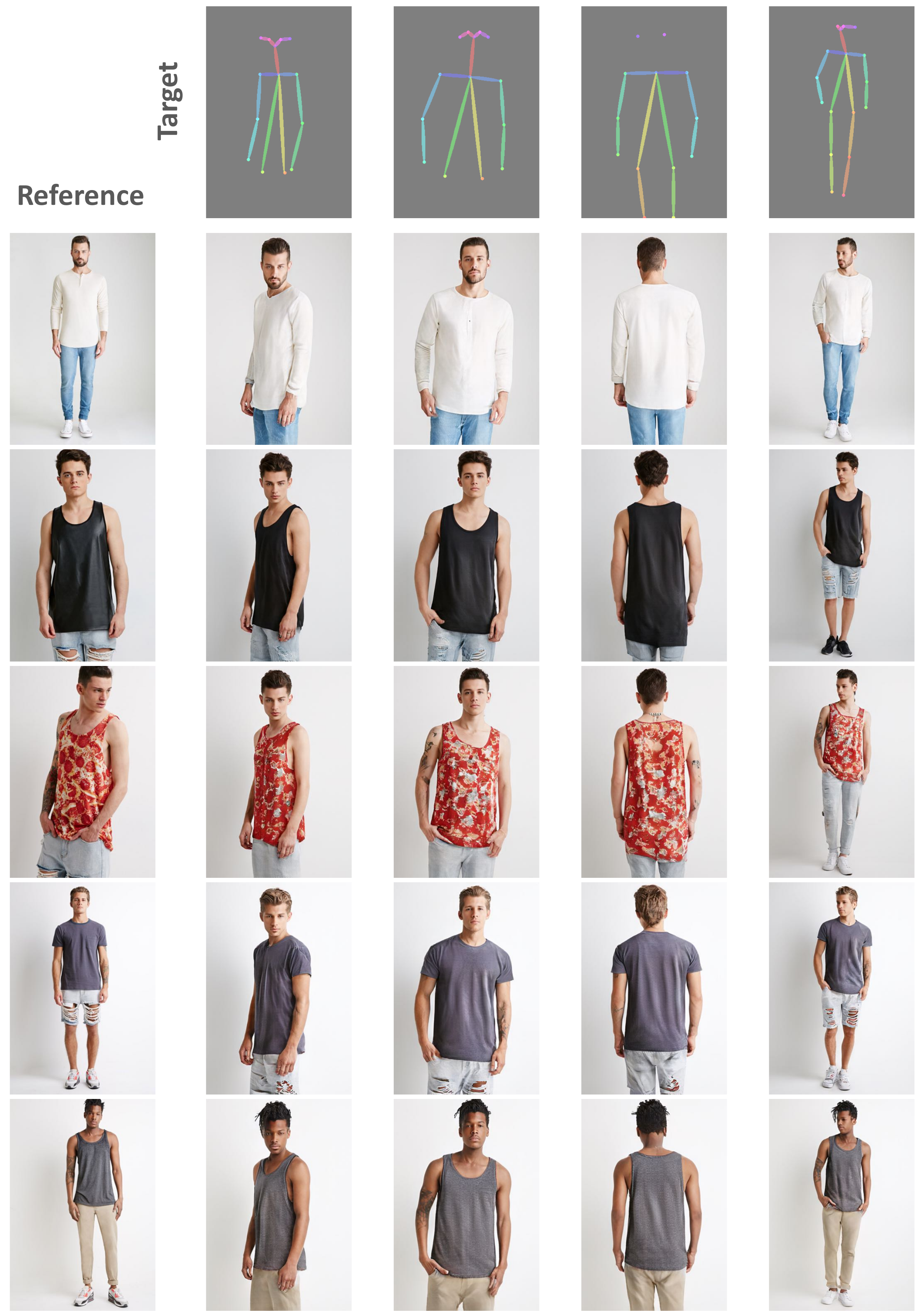}
\end{center}
   \caption{Additional results of pose control.}

\label{fig:pose_control_0}
\end{figure*}

\begin{figure*}[ht]
\renewcommand\thefigure{\Alph{alphasect}.\arabic{figure}}
\begin{center}
\includegraphics[width=0.81\linewidth]{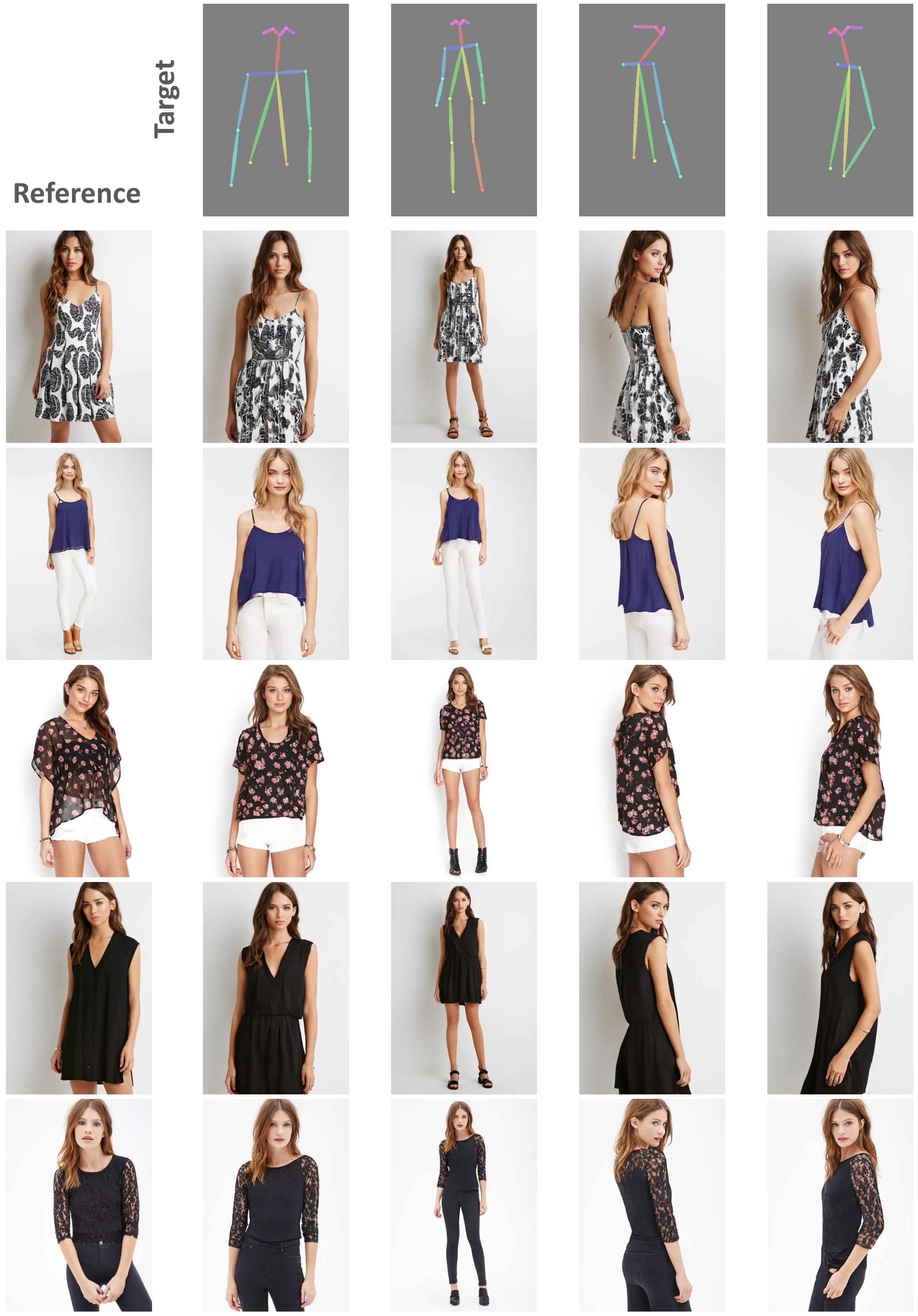}
\end{center}
   \caption{Additional results of pose control.}

\label{fig:pose_control_1}
\end{figure*}

\begin{figure*}[ht]
\renewcommand\thefigure{\Alph{alphasect}.\arabic{figure}}
\begin{center}
\includegraphics[width=0.81\linewidth]{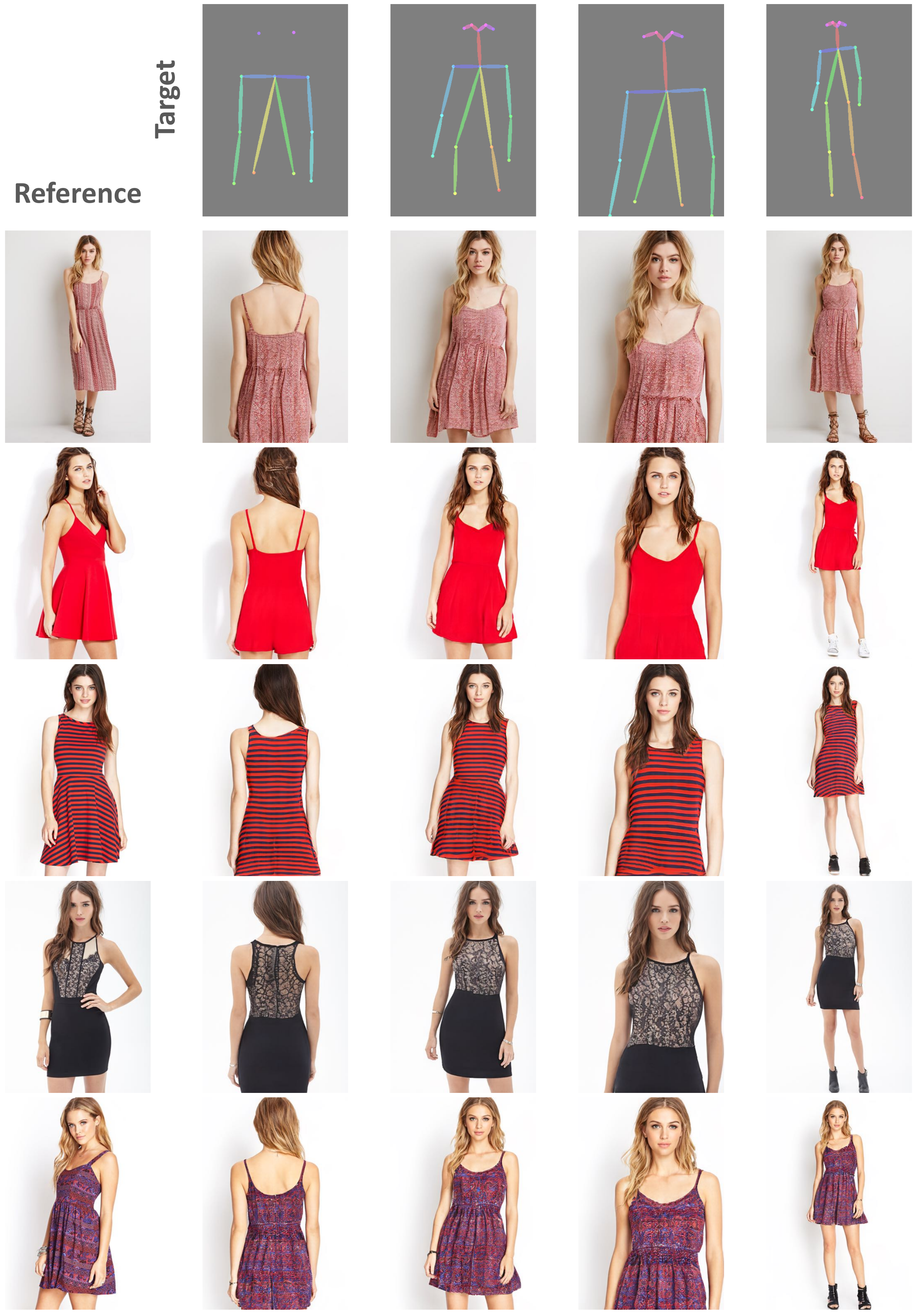}
\end{center}
   \caption{Additional results of pose control.}

\label{fig:pose_control_2}
\end{figure*}

\begin{figure*}[ht]
\renewcommand\thefigure{\Alph{alphasect}.\arabic{figure}}
\begin{center}
\includegraphics[width=0.81\linewidth]{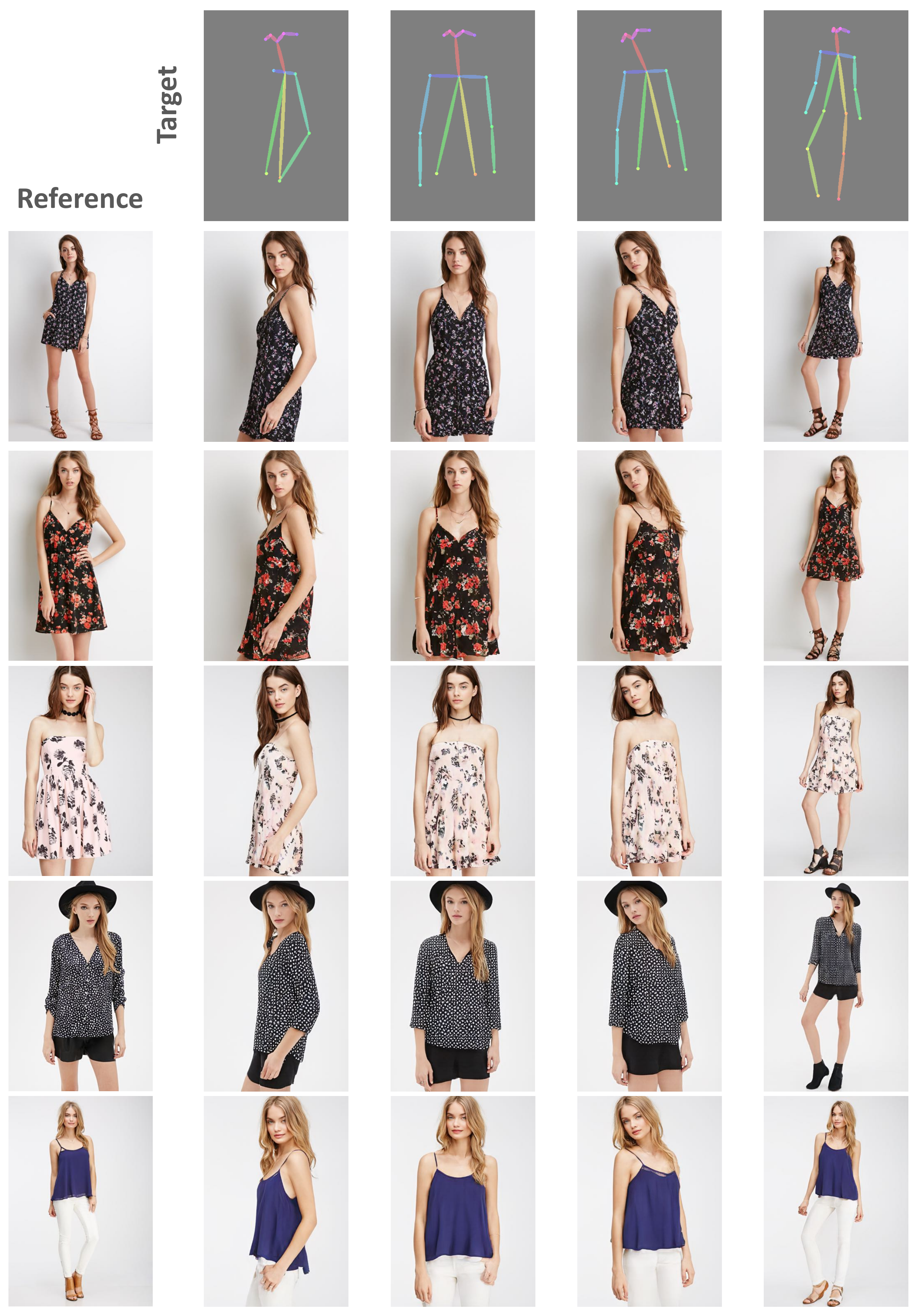}
\end{center}
   \caption{Additional results of pose control.}

\label{fig:pose_control_3}
\end{figure*}

\subsection{Additional Results of Appearance Control} 
We provide additional results of appearance control in Fig.~\ref{fig:appearance_control_0}, Fig.~\ref{fig:appearance_control_1}, and Fig.~\ref{fig:appearance_control_2}.

\begin{figure*}[h]
\renewcommand\thefigure{\Alph{alphasect}.\arabic{figure}}
\begin{center}
\includegraphics[width=0.9\linewidth]{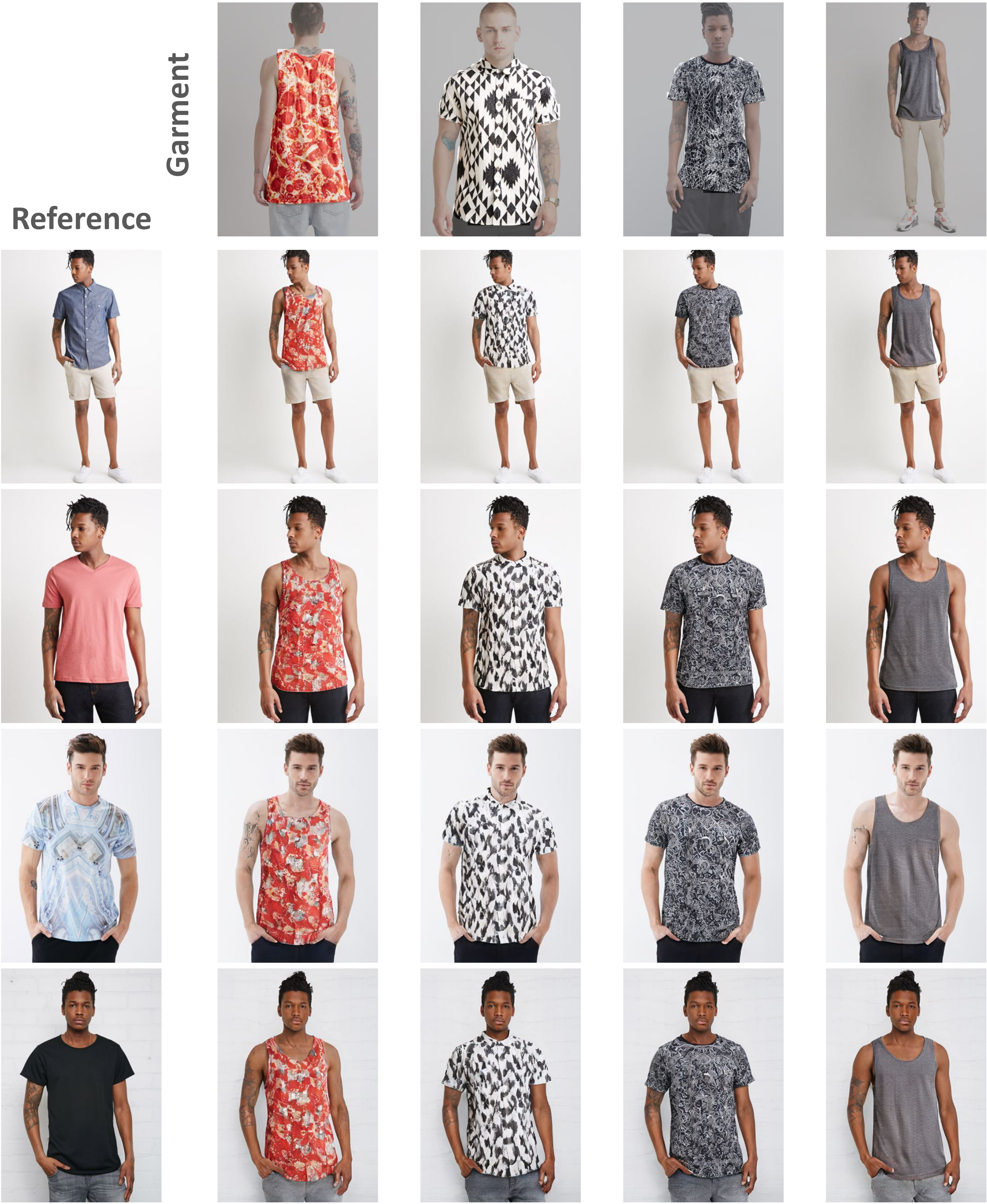}
\end{center}
   \caption{Additional results of appearance control.}

\label{fig:appearance_control_0}
\end{figure*}

\begin{figure*}[h]
\renewcommand\thefigure{\Alph{alphasect}.\arabic{figure}}
\begin{center}
\includegraphics[width=0.9\linewidth]{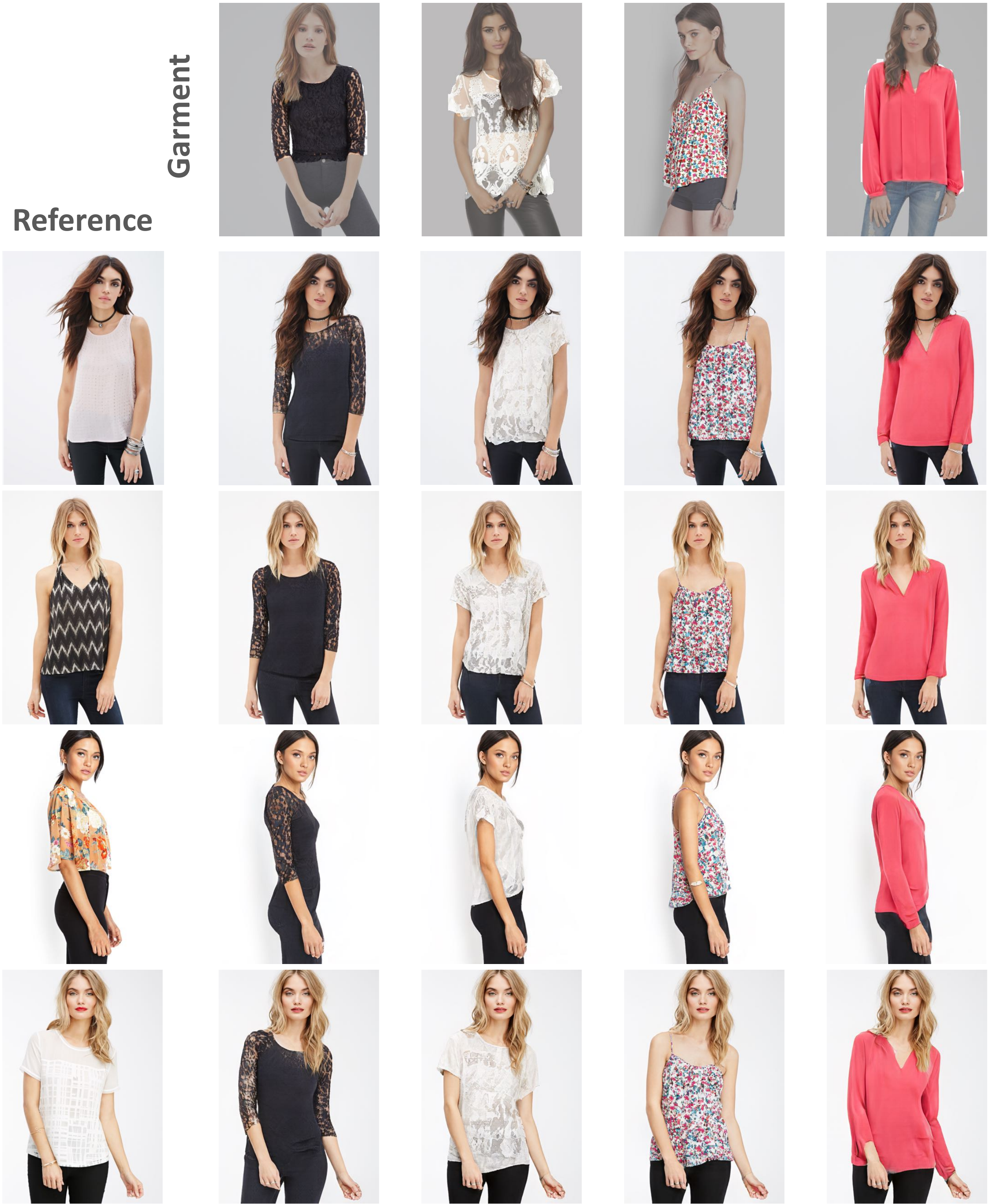}
\end{center}
   \caption{Additional results of appearance control.}

\label{fig:appearance_control_1}
\end{figure*}

\begin{figure*}[h]
\renewcommand\thefigure{\Alph{alphasect}.\arabic{figure}}
\begin{center}
\includegraphics[width=0.9\linewidth]{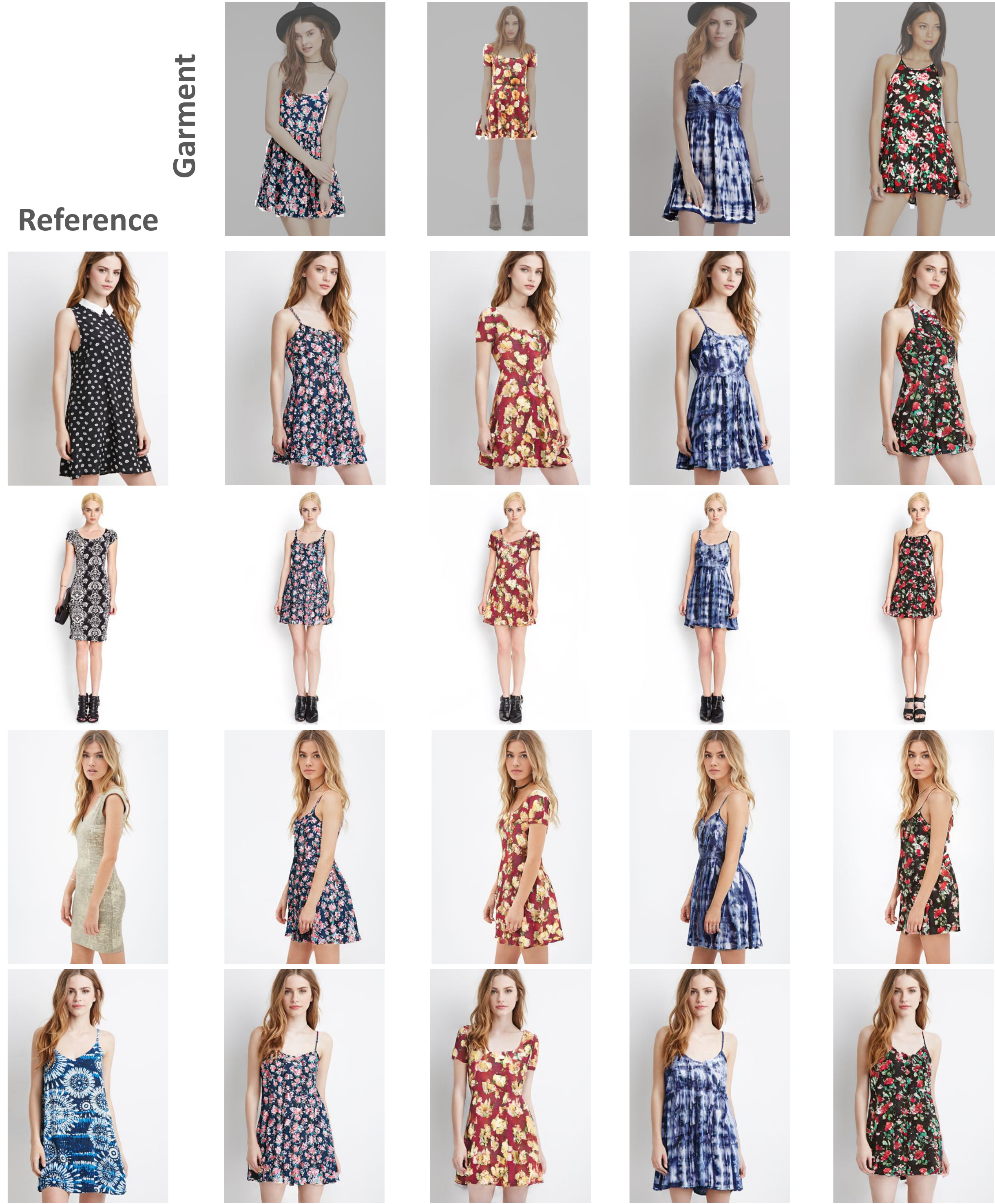}
\end{center}
   \caption{Additional results of appearance control.}

\label{fig:appearance_control_2}
\end{figure*}

\clearpage

\end{alphasection}
\end{document}